\documentclass{article}
\usepackage{amssymb}
\usepackage{graphicx}
\usepackage{booktabs}
\usepackage{makecell}  
\usepackage{fontawesome5}
\usepackage{amsmath} 
\usepackage{hyperref}
\usepackage{wrapfig} 
\usepackage{mathrsfs}
\usepackage{pifont} 
\usepackage{multirow}   
\usepackage{arydshln}   
\usepackage{paralist}
\usepackage{enumitem}
\usepackage{caption}
\usepackage[table,dvipsnames]{xcolor}
\usepackage{colortbl}
\newcommand{\redx}{\textcolor{red}{\times}}
\newcommand{\greencheck}{\textcolor{green!60!black}{\checkmark}}

\newcommand{\yellowtri}{\textcolor{yellow!90!black}{\blacktriangle}}
\usepackage[preprint]{corl_2026} 

\title{TBD-VLA: Temporal Block Diffusion  \\
Vision Language Action Model
}

\author{
  Sung-Wook Lee, Xuhui Kang, Yen-Ling Kuo\\
  University of Virginia\\
  \texttt{\{dcs3zc,xuhui,ylkuo\}@virginia.edu}
  \\
}

\begin{document}
\maketitle

\begin{abstract}
Discrete Vision-Language-Action (VLA) models typically formulate action generation as next-token prediction over discretized action spaces, conditioning each token autoregressively on prior context. While effective, this paradigm incurs high inference latency and largely ignores the temporal structure inherent in action trajectories. Recent efforts introduce parallel decoding to improve efficiency, enabling faster inference, but lack explicit mechanisms for modeling token dependencies.
We introduce TBD-VLA, a discrete token-based VLA framework that incorporates block diffusion to enable temporal action generation. We partition action sequences into temporal blocks and perform masked discrete diffusion within each block, while maintaining autoregressive generation across blocks. This design unifies temporal autoregression and parallel action decoding, achieving both strong temporal coherence and improved inference speed.
In addition, the explicit temporal modeling enables asynchronous execution of action chunks (e.g., Real-Time Chunking) via temporal in-painting. TBD-VLA significantly outperforms prior VLA approaches in both simulation and real-world manipulation tasks, offering a scalable path toward fast, temporally aware, discrete VLA models.

\end{abstract}


\section{Introduction}

Vision-Language-Action (VLA) models have emerged as a promising paradigm for building generalist robotic policies, leveraging large-scale pretraining to map visual observations and natural language instructions into executable robot actions. A central design question in this space is how a vision-language model (VLM) backbone contributes to action generation, and the field has converged on several distinct answers, each with its own trade-offs. The currently dominant approach, exemplified by $\pi_{0.5}$ \cite{pmlr-v305-black25a} and GR00T N1.5 \cite{bjorck2025gr00t}, attaches a continuous action expert, typically a flow-matching head on top of the VLM backbones, which naturally handles the continuous and multimodal action sequences. However, decoupling the VLM from action generation makes it fundamentally harder to analyze what the VLM exactly contributes to VLA's capability to generalize.

An alternative is to use the VLM itself as the action decoder by representing actions as discrete tokens that can be directly generated by the model. While promising, a significant challenge lies in the efficiency: autoregressive generation of long action chunks, one token at a time, is prohibitively slow for closed-loop, high-frequency robot control. Recent efforts to make token-based action decoding practical have largely followed two complementary directions. One direction focuses on improving the representation of action tokens: instead of directly tokenizing dense timestep-wise action sequences, actions can be transformed into more compact or structured representations, reducing the number of tokens the VLM must generate \cite{pertsch2025fast, wang2025vq, liu2026oat}. While this improves efficiency, such representations may weaken the explicit correspondence between individual tokens and localized timesteps. Another direction focuses on improving the decoding procedure itself by generating multiple action tokens in parallel rather than strictly autoregressively \cite{kim2025fine, liang2025discrete}. 
This can substantially reduce inference latency, but existing parallel decoding approaches still provide limited mechanisms for modeling temporal dependencies across actions.

To address these limitations, we introduce Temporal Block Diffusion Vision-Language-Action (TBD-VLA), a discrete token-based VLA framework that formulates action generation as blockwise discrete diffusion \cite{arriola2025block, wu2025fast}. TBD-VLA partitions action sequences into temporal blocks, decoding tokens in parallel within each block while generating blocks autoregressively. This design combines the efficiency of parallel decoding with explicit temporal-level autoregression, enabling temporally coherent action generation and faster inference. Furthermore, its temporal modeling enables Real-Time Chunking (RTC) \cite{black2026real}, an asynchronous inference mechanism that mitigates inference latency: Since TBD-VLA naturally incorporates inpainting (unmasking) during training, the model is aligned to complete partially committed action chunks. Therefore, this training-inference alignment lead to superior performance when compared to the baseline methods.

Our contributions are as follows:
1) We introduce TBD-VLA, a framework for Vision Language Action model that combines parallel action decoding and temporal-level autoregression.
2) We develop a novel scheme for incorporating block discrete diffusion into efficient VLA training pipeline.
3) We perform extensive evaluations on multiple benchmarks in simulation and in real-world under various perturbation scenarios and show a strong generalizable manipulation capability of our model.


\begin{figure*}[t]
    \begin{center}
        {\includegraphics[width=\linewidth]{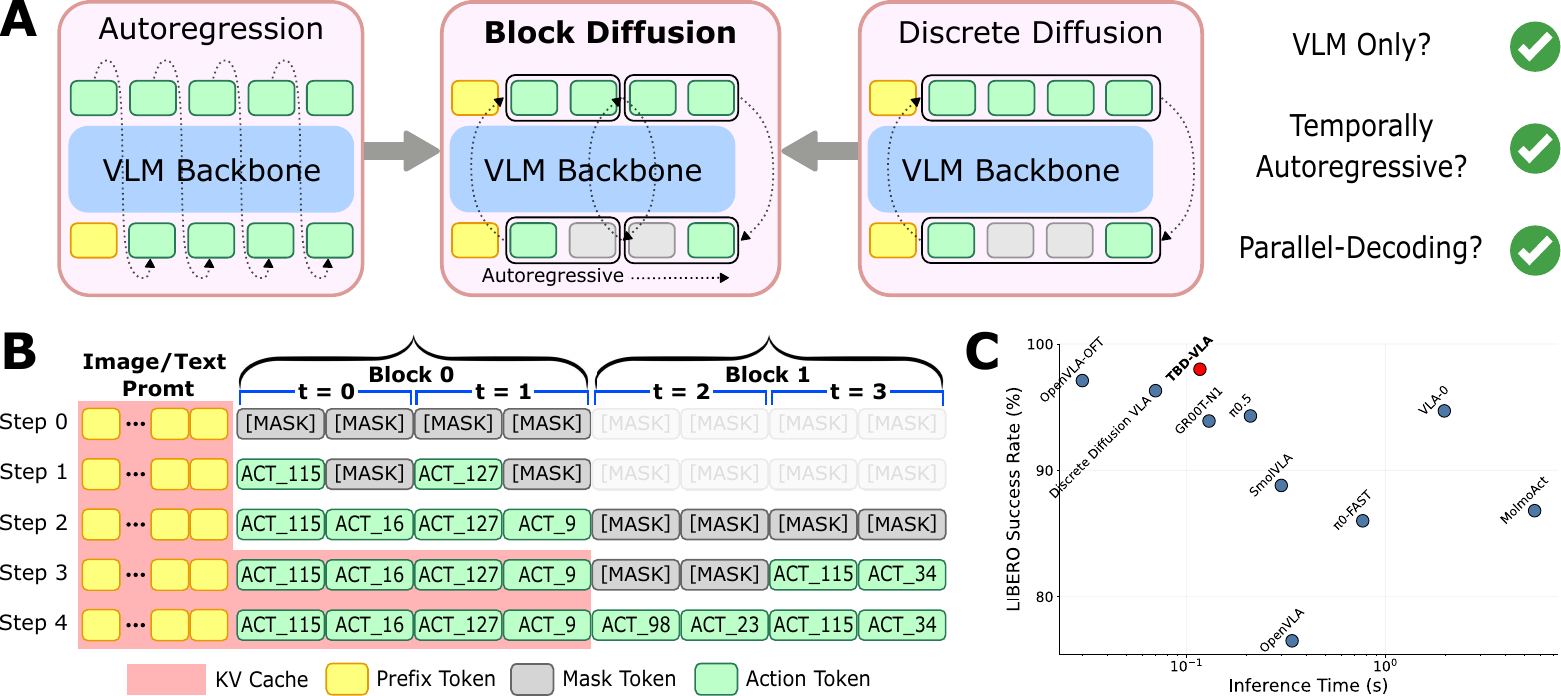} 
        }
        
\caption{\textbf{Overview of Temporal Block Diffusion Vision Language Action (TBD-VLA) model.} 
\textbf{(A)\:}TBD-VLA formulates action sequence generation as block discrete diffusion, which incorporates autoregression and discrete diffusion into a single framework.
\textbf{(B)\:}At inference time, action tokens are decoded in parallel within blocks and autoregressively between blocks. KV caching for prefix further accelerates inference. 
\textbf{(C)} TBD-VLA achieves the SOTA results on multiple benchmarks in simulation and in real-world while retaining a competitive inference speed.}
\vspace{-12pt}
\label{fig:main}
\end{center}
\end{figure*}
\section{Related Work}
\textbf{Masked Diffusion.}
Masked diffusion models~\cite{sohl2015deep, nie2026large, sahoo2024simple} generate discrete sequences through iterative masked-token prediction, enabling many tokens to be refined in parallel rather than decoded strictly left-to-right. 
This paradigm has been applied to multimodal generation to improve sampling efficiency while retaining expressive token-level dependencies~\cite{swerdlow2025unified, yang2026mmada}. 
Recent block diffusion models combine parallel denoising within blocks with autoregressive generation across blocks, providing an efficient compromise between parallel decoding and causal sequence modeling~\cite{arriola2025block, wu2025fast}. 
TBD-VLA brings this blockwise masked-diffusion formulation to action generation, using temporal action blocks as the unit of autoregression.

\textbf{Discrete Vision Language Action Models.}
Discrete VLA frameworks have recently emerged as a promising direction for enabling VLMs to decode robot actions directly. Recent work addresses the efficiency bottleneck of discrete VLA frameworks through either more compact action representations or faster decoding. For example, \citet{pertsch2025fast} compresses action trajectories into frequency-domain tokens, while learning-based methods \cite{wang2025vq, liu2026oat} learn discrete latent action vocabularies to reduce the token sequence length. While these existing methods improve decoding efficiency, they lack the temporal modeling capability. Other methods aim to accelerate decoding itself: OpenVLA-OFT \cite{kim2025fine} improves OpenVLA \cite{kim2024openvla} with fully parallel action decoding, and discrete diffusion-based VLAs \cite{song2026fast, chen2025unified, liang2025discrete, wen2025llada} replace left-to-right generation with iterative masked-token refinement, enabling multi-step parallel decoding. However, existing parallel or diffusion-based decoders typically provide limited explicit modeling of temporal dependencies across action chunks.

TBD-VLA builds on discrete-diffusion VLAs, but introduces temporal block structure into action generation. It performs masked discrete diffusion within each temporal block while generating blocks autoregressively, combining within-block parallel decoding with explicit temporal dependency modeling. Unlike compressed-token methods, TBD-VLA preserves timestep-level action tokens; unlike other parallel decoding methods, it retains temporal autoregression across blocks. 

\begin{table}[t]
\centering
\small
\resizebox{0.99\textwidth}{!}{%
\begin{tabular}{l c c c c}
\toprule
\textbf{Model Name} & \textbf{Model Size} & \textbf{Temporal AR} & \textbf{Action Decoder} & \textbf{Latency (s) $\downarrow$} \\
\midrule
SmolVLA \;\cite{shukor2025smolvla} & 0.5B & $\redx$ & Flow Matching & 0.297 \\
GR00T-N1 \;\cite{bjorck2025gr00t} & 2.2B & $\redx$ & Flow Matching & 0.131 \\
$\pi_{0.5}$ \;\cite{pmlr-v305-black25a} & 3B & $\redx$ & Flow Matching & 0.208 \\
\midrule
OpenVLA \;\cite{kim2024openvla} & 7B & $\redx$ & Autoregressive & 0.344 \\
OpenVLA-OFT \;\cite{kim2025fine} & 7B & $\redx$ & Parallel & \textbf{0.031} \\
MolmoAct \;\cite{lee2025molmoact} & 7B & $\redx$ & Autoregressive & 5.633 \\
$\pi_0$-FAST \;\cite{pertsch2025fast} & 3B & $\redx$ & Autoregressive & 0.767 \\
Discrete Diffusion VLA \;\cite{liang2025discrete} & 7B & $\redx$ & Discrete Diffusion & 0.069 \\
VLA-0 \;\cite{goyal2025vla} & 3B & $\yellowtri$ & Autoregressive & 1.980 \\
\textbf{TBD-VLA} & 2B & $\greencheck$ & Block Discrete Diffusion & 0.117 \\
\bottomrule
\end{tabular}
}
\vspace{3pt}
\caption{Comparison of VLA models by model size, temporal autoregression (AR), action decoding strategy, and action generation latency in LIBERO environment. Note that VLA-0 is autoregressive in text strings.}
\vspace{-10pt}
\label{tab:vla_comparison}
\end{table}

\section{Problem Statement}
We consider visuomotor policy learning in a vision--language setting, where the goal is to learn a policy $\pi_\theta(a_{1:H} \mid o, g)$ that maps an observation $o$, consisting of visual inputs and proprioceptive state, and a task specification $g$ (e.g., language), to a sequence of future robot actions $a_{1:H_p}$ where $H_p$ is the action prediction horizon.
To enable the use of vision--language models for control directly, we represent actions as discrete tokens: Let $A_t = [a_t, \dots, a_{t+H_{p-1}}]$ denote an action chunk, and each action feature is discretized into $N_b$ bins. Each discretized feature corresponds to a token drawn from a vocabulary $\mathcal{V}$ of size $|\mathcal{V}| = N_b$. Thus, an action chunk $A_t$ is represented as a sequence of tokens of length $L_{t} = H_p \cdot D_a$, where $D_a$ is action dimension. 
We focus on temporally autoregressive action generation, where the action sequence likelihood is factorized over temporal action blocks as
\[
\textstyle
p(a_{1:H_p}\mid o,g)
=\prod_{k=0}^{K-1}
p_\theta(a_{km+1:(k+1)m}\mid o,g,a_{1:km}),
\]
where \(m\) denotes the temporal size of action block and \(K = H_p/m\) denotes the number of blocks.


\section{Method}
\subsection{Model Architecture}
\paragraph{Base Model and Tokenization} We use Qwen3-VL 2B \cite{bai2025qwen3} as the VLM backbone, although our method is compatible with any VLM backbones. 
We augment the VLM tokenizer with special tokens, including mask tokens, placeholder tokens, and action tokens. Both proprioception and action feature is discretized into $N_b$ bins and tokenized using the shared dictionary. The VLM is prompted with the following template: ``State: \{state tokens\}, Task: \{instruction\}, Actions: \{placeholder tokens\}", where the placeholder tokens guide how many action tokens to generate.

\subsection{Training Pipeline}

\begin{figure*}[t]    
    \begin{center}
        {\includegraphics[width=\linewidth]{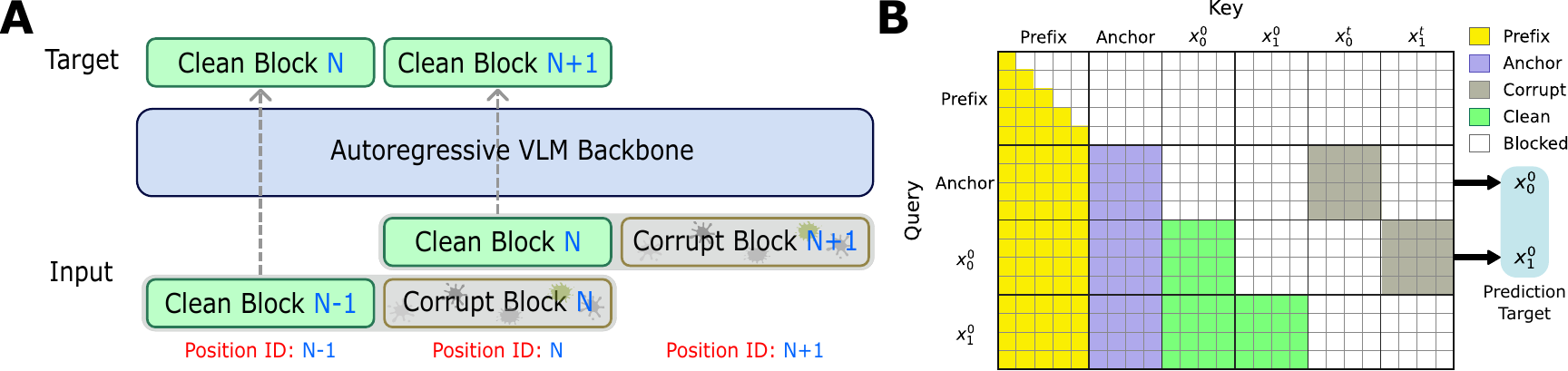}} 
        
\caption{
\textbf{Training for TBD-VLA.
} 
\textbf{(A)} In order to match the VLM backbone's autoregressive property, we apply token shift, where the logits for the current action block are generated from the prior block. \textbf{(B)} A doubled-layout trick is used, in which clean and partially masked (corrupt) action blocks are processed in parallel under a custom attention mask.
}
\vspace{-10pt}
\label{fig:train}
\end{center}
\end{figure*}

\paragraph{Temporal-level Token Shift} 
To better align with the pretrained VLM backbone's next-token prediction objective, we shift the prediction target at the temporal level, where the tokens from the current action block are trained to predict the next action block. This design bridges the gap between the self-reconstructive formulation of discrete diffusion and the next-token prediction of the autoregressive VLM backbone. See Figure. \ref{fig:train} (A) for visualization of the temporal-level token shift.

\paragraph{Discrete Block Diffusion}

We model action generation with block-wise discrete diffusion. 
Let \(x^0=\tau(a_{1:H})\) be the tokenized action sequence, where \(\tau\) denotes the action tokenizer. We partition \(x^0\) into \(K\) blocks, \(x^0=(x_0^0,\ldots,x_{K-1}^0)\), where \(x_m^0 \in \mathcal{V}^{m \cdot D_a}\).
During the forward process, we construct a corrupted block \(x_k^t\) from the corresponding clean block \(x_k^0\), where superscripts \(0\) and \(t\) denote the clean and corrupted action blocks, respectively. For token position \(i\) within each block \(k\), we sample \(t_{k,i}\sim\mathcal{U}(0,1)\) as the masking probability:
During the forward process, each clean block \(x_k^0\) is corrupted into \(x_k^t\) by independently masking each token with \(t_{k,i}\sim\mathcal{U}(0,1)\). The reverse process predicts the clean tokens of block \(k\) conditioned on a shifted predictor block \(z_k\), which uses the anchor block for the first action block and otherwise contains the clean preceding blocks:
\begin{equation}
\begin{minipage}{0.4\linewidth}
\centering
\(\displaystyle
x_{k,i}^{t}\sim
\begin{cases}
\texttt{[MASK]}, & \Pr=t_{k,i},\\
x_{k,i}^{0}, & \Pr=1-t_{k,i},
\end{cases}
\)
\end{minipage}
\hfill
\begin{minipage}{0.37\linewidth}
\centering
\(\displaystyle
z_k =
\begin{cases}
s, & k=0,\\
x_{0:k-1}^0, & k>0,
\end{cases}
\)
\end{minipage}
\end{equation}
where \(s=(\texttt{[MASK]},\ldots,\texttt{[MASK]})\) denotes the anchor block.
where $z_k$ contains the clean preceding blocks when $k>0$, and $s=(\texttt{[MASK]},\ldots,\texttt{[MASK]})$ is an anchor block used when predicting the first action block.
The loss is the average cross-entropy over masked action tokens:
\[
\mathcal{L}_\theta
=
-
\frac{
\sum_{k=0}^{K-1}
\sum_{i=0}^{m \cdot D_a-1}
\mathbf{1}[x_{k,i}^{t}=\texttt{[MASK]}]\,
\log p_\theta(x_{k,i}^0 \mid z_k, x_{k}^t, o, g)
}{
\sum_{k=0}^{K-1}
\sum_{i=0}^{m \cdot D_a-1}
\mathbf{1}[x_{k,i}^{t}=\texttt{[MASK]}]
}.
\]

\paragraph{Block-level Attention Masking}
To enable efficient training for block masked diffusion, where both intra-block parallelism and inter-block autoregression must be handled at once, we use a custom attention mask similar to \cite{arriola2025block,wu2025fast}: We use a doubled-layout trick, where the clean action sequence $x^0$ and noised sequence $x^t$ are concatenated as inputs while sharing the same RoPE positions. To predict $n$-th clean action block $x^0_{n}$, the policy is given context of the prefix $(o, g)$, the previous action blocks $x^0_{0:n-1}$, and the $n$-th corrupt action block $x^t_{n}$. The custom attention map parallelizes learning across multiple action blocks in a single pass, significantly accelerating the training efficiency. See Figure. \ref{fig:train} (B) for visualization of the attention map. 

\subsection{Inference} 

\paragraph{Decoding as Needed}
At inference time, we generate action blocks sequentially from fully masked
tokens. Each decoded block is refined for $n_d$ discrete diffusion steps, where
at each step the model predicts all masked positions and commits the most
confident tokens. To reduce latency, the policy decodes only the blocks needed
for execution: for rollout horizon $H_{\mathrm{a}}$, it generates
$K_{\mathrm{exec}}=\lceil H_{\mathrm{a}}/m\rceil$ blocks instead of all
$K=H_p/m$ blocks, requiring $K_{\mathrm{exec}} \cdot n_d$ denoising steps in total.

\paragraph{Prefix KV Cache}
To improve inference efficiency, TBD-VLA caches the key–value states of the current visual and prompt tokens, as well as those of previously generated action blocks. During the discrete diffusion process, KV caching avoids redundant computation of this unchanged context across denoising steps. 

\paragraph{Action Decoding}
Action tokens are unmasked in order of confidence, with higher-logit tokens decoded first. 
For the action chunk \( A_t=[ a_t,\ldots, a_{t+H_p-1}]\), 
we propose expectation sampling to decode each scalar action component from the full predicted token distribution. 
Specifically, for timestep \(h\) and action dimension \(j\), the scalar action value is decoded as
\( a_{t+h,j}=\sum_{x\in\mathcal{V}} p_{\theta,h,j}(x)c_j(x)\),
where \(p_{\theta,h,j}(x)\) is the predicted probability of action token \(x\in\mathcal{V}\), and \(c_j(x)\) maps the token to the raw action value of the corresponding bin for action dimension \(j\). 
This uses the complete output distribution as a finer-grained signal instead of the most likely discrete token.

\paragraph{Real-Time Chunking}
To mitigate inference latency during closed-loop control, we support Real-Time Chunking (RTC), which asynchronously generates future actions while executing the current actions. Specifically, we adopt a hard in-painting strategy, in which the previously generated action tail corresponding to the inference-latency window is frozen and reused as in-painting context for the early action blocks. This aligns with TBD-VLA's masked block-diffusion objective, which trains the model to complete action blocks conditioned on partial action context.

\begin{figure*}[t]
    \begin{center}
        {\includegraphics[width=0.8\linewidth]{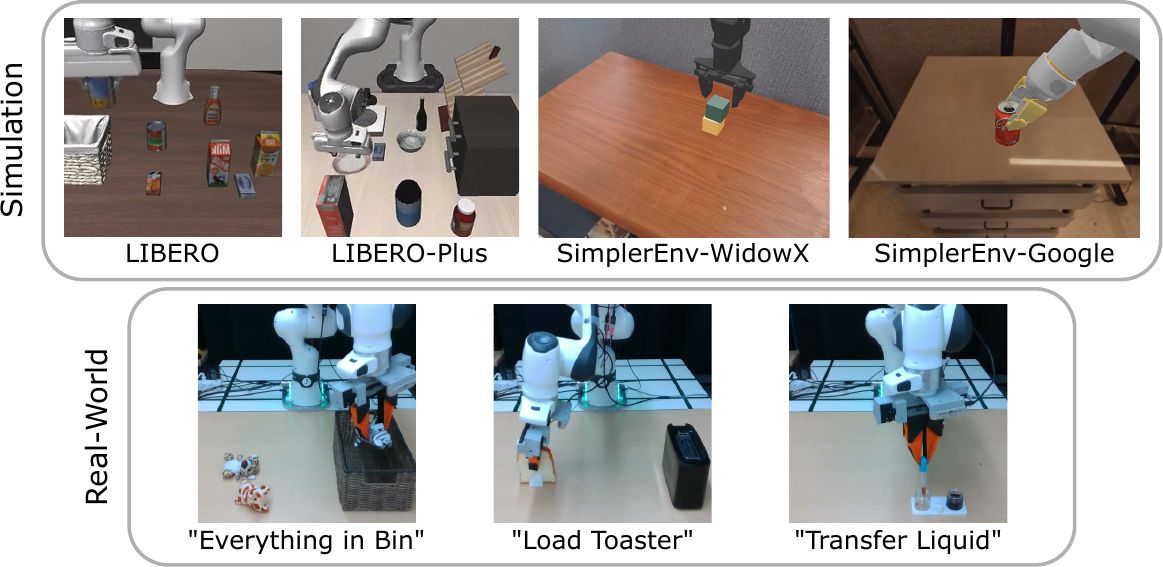} 
        }    
\caption{
\textbf{Benchmarks and tasks.} In simulation, TBD-VLA is evaluated across multiple robots: LIBERO and LIBERO-Plus using a Franka Panda robot arm, and SimplerEnv using the Google Robot and Widow-X arm. In real-world, three tabletop tasks are used to evaluate with a Franka Research 3 (FR3) arm.
}
\vspace{-10pt}
    \end{center}
    \label{fig:sim}%
\end{figure*}

\section{Experiments}

We conduct extensive experiments in both simulation and real-world. Our investigation addresses the following research questions.
\begin{compactenum}
    \item[\textbf{RQ1}] How effectively does TBD-VLA generalize across diverse evaluation settings, including multiple robotic platforms and varying perturbation scenarios?
    \item[\textbf{RQ2}] Does incorporating RTC improve TBD-VLA’s performance on real-world tasks?
    \item[\textbf{RQ3}] Which design choices contribute to the model’s performance and inference speed?
\end{compactenum}

\subsection{Benchmarks}

\paragraph{LIBERO}
LIBERO~\citep{liu2023libero} is a manipulation benchmark comprising four suites: Spatial, Object, Goal, and Long, which evaluate spatial reasoning, object generalization, goal conditioning, and long-horizon task execution, respectively. We report success rates for each suite and the overall average, with 10 tasks per suite and 50 rollouts per task. In addition, we test TBD-VLA with RTC under inference latency, simulated as delayed observations in simulation steps.

\vspace{-4pt}
\paragraph{LIBERO-Plus}
LIBERO-Plus~\citep{fei2025libero} extends LIBERO with controlled perturbations for robustness evaluation. It tests policies under variations in object layout, camera viewpoint, robot initial state, language instruction, lighting, background texture, and sensor noise. We train the model on the original LIBERO datasets and report the zero-shot success rates under perturbations across 10,030 rollouts.

\vspace{-4pt}
\paragraph{SimplerEnv}
SimplerEnv~\citep{li2024evaluating} is a real-to-sim benchmark for evaluating the transfer and generalization of robot policies trained on real-world data. We evaluate TBD-VLA on pre-defined Widow-X tasks and Google Robot tasks under visually matching and visually aggregated settings. We report per-task success rates and the overall average for final success.

\begin{table}[t]
\centering

\begin{minipage}[t]{0.54\textwidth}
\vspace{-2pt}
\centering
\setlength{\tabcolsep}{5.7pt}
\renewcommand{\arraystretch}{1.35}

\resizebox{\linewidth}{!}{%
\begin{tabular}{lccccc}
\toprule
Model & Spatial & Object & Goal & Long & Avg \\
\midrule
OpenVLA-oft \;\cite{kim2025fine}       & 96.2 & 98.3 & 96.2 & 90.7 & 95.4 \\
$\pi_{0}$-Fast \;\cite{pertsch2025fast}    & 96.4 & 96.8 & 88.6 & 60.2 & 85.5 \\
$\pi_{0.5}$ \;\cite{pmlr-v305-black25a}       & \textbf{98.8} & 98.2 & \underline{98.0} & 92.4 & \underline{96.9} \\
GR00T-N1 \;\cite{bjorck2025gr00t}          & 94.4 & 97.6 & 93.0 & 90.6 & 93.9 \\
MolmoAct \;\cite{lee2025molmoact}          & 87.0 & 95.4 & 87.6 & 77.2 & 86.6 \\
UniVLA \;\cite{wang2025unified}            & 95.4 & \underline{98.8} & 93.6 & \underline{94.0} & 95.5 \\
VLA-0 \;\cite{goyal2025vla}             & 97.0 & 97.8 & 96.2 & 87.6 & 94.7 \\
Disc Diff VLA \;\cite{liang2025discrete}     & 97.2 & 98.6 & 97.4 & 92.0 & 96.3 \\
UD-VLA \;\cite{chen2025unified}            & 94.1 & 95.7 & 91.2 & 89.6 & 92.7 \\
dVLA \;\cite{song2026fast}            &  97.4 & 97.9 & \textbf{98.2} & 92.2 & 96.4 \\
\midrule
\textbf{TBD-VLA}  & \underline{97.6} & \textbf{99.6} & 97.4 & \textbf{96.6} & \textbf{97.7} \\
\bottomrule
\end{tabular}%
}
\end{minipage}%
\hspace{0.02\textwidth}%
\begin{minipage}[t]{0.44\textwidth}
\vspace{-4pt}
\centering
\includegraphics[width=\linewidth]{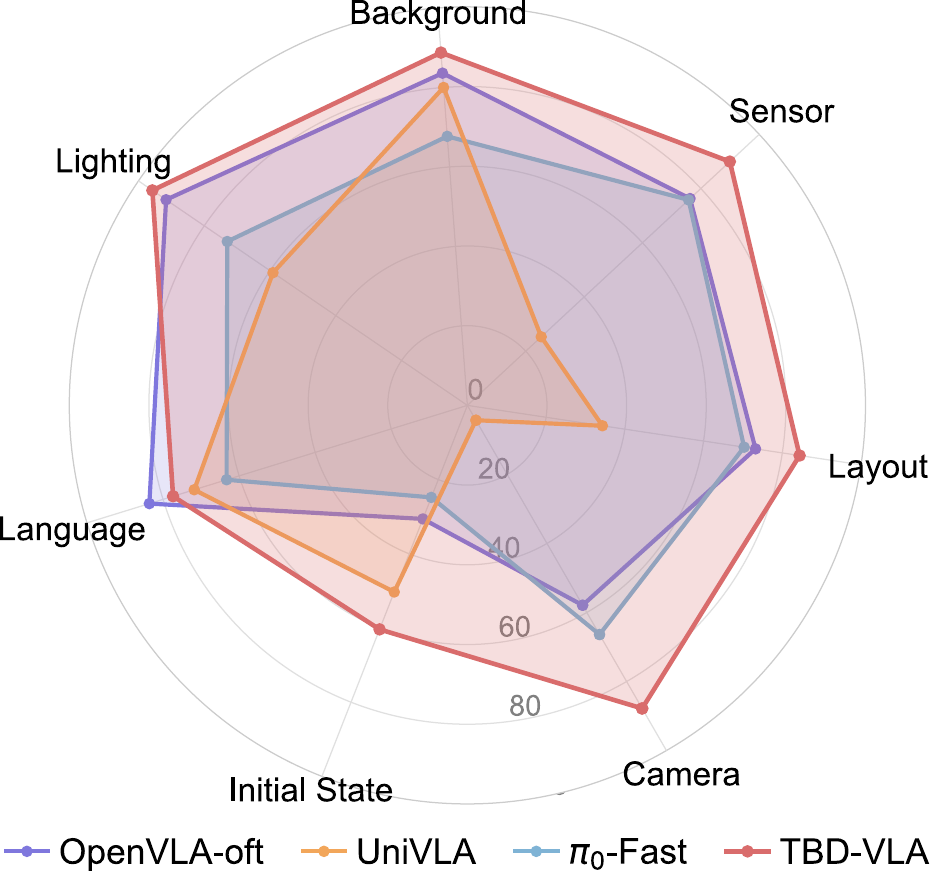}
\end{minipage}
\vspace{5pt}
\caption{Left: Success rates (\%) on the LIBERO benchmark across the four task suites. Best result per column in \textbf{bold}; second-best \underline{underlined}. Right: Zero-shot success rates (\%) on LIBERO-Plus for each of the perturbation scenarios across the four task suites. 
}
\vspace{-10pt}
\label{fig:libero_combined}
\end{table}

\subsection{Pre-training and Fine-tuning} 
In all experiments, the VLM backbones are pre-trained on large-scale, open-source community datasets, including DROID \cite{khazatsky2024droid}, Open-X Embodiment \cite{o2024open}, RoboSet \cite{kumar2023robohive}, RoboMIND \cite{wu2024robomind}, and RH20T \cite{fang2023rh20t}. 
For SimplerEnv Widow-X benchmark, the policy is fine-tuned on Bridge-V2 dataset \cite{walke2023bridgedata} for 20K training steps. For SimplerEnv Google Robot benchmark, it is fine-tuned on Fractal dataset \cite{brohan2022rt} for 40K steps. For LIBERO and LIBERO-Plus benchmarks, the policy is fine-tuned on the single, original LIBERO task suites dataset for 80K steps. For all cases, the temporal block size $m$ is set as 4 and the prediction horizon $H_p$ is set as 16. For training details, see Appendix \ref{app:train}.

\subsection{Simulation Results}

\begin{wrapfigure}[15]{r}{0.44\linewidth}
    \centering
    \vspace{-2pt}
        \begin{minipage}{0.98\linewidth}
            \centering
            \includegraphics[width=\linewidth]{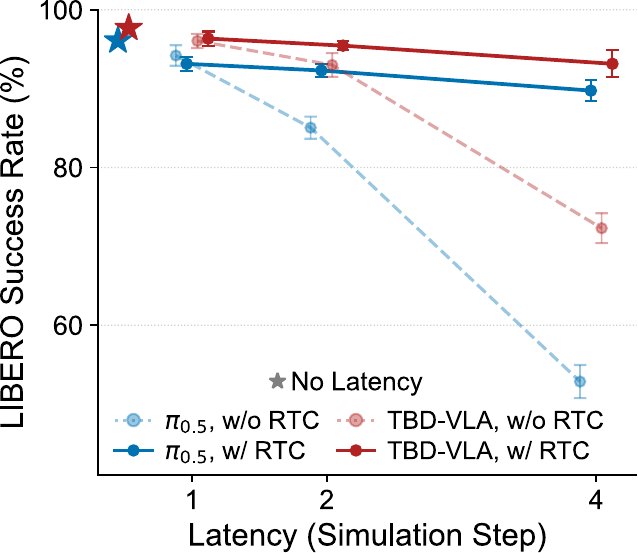}
            \caption{LIBERO success rate with/without RTC vs. latency. Stars denote zero added latency.}
            \label{fig:rtc}
        \end{minipage}
\end{wrapfigure}

\paragraph{LIBERO and LIBERO-Plus} Table~\ref{fig:libero_combined} summarizes the simulation performance of TBD-VLA compared to other models. 
TBD-VLA achieves the SOTA results on the LIBERO test suites at 97.6\% average success rate. As shown in Figure \ref{fig:rtc}, under the inference delay of 4 simulation steps, TBD-VLA with RTC retains 93.2\% success rate, which is 3.4\% higher than $\pi_{0.5}$ with RTC. 
Notably, the policy performance for TBD-VLA without RTC degrades to 72.3\% under the same latency, showing the effectiveness of the asynchronous inference. 
Furthermore, TBD-VLA shows high robustness against various perturbation evaluations in LIBERO-Plus, achieving 83.0\% success rate on average.
outperforming the second best method by 15.1\%. 
For full results, refer to Appendix~\ref{app:libero_results}.

\begin{table}[b!]
\centering
\caption{Success rates (\%) on the SimplerEnv Widow-X benchmark. ``Avg" indicates the average score for the final success rate.}
\vspace{2pt}
\small
\renewcommand{\arraystretch}{1.2}
\resizebox{0.97\textwidth}{!}{%
\begin{tabular}{l|cc cc cc cc c}
\toprule
\multicolumn{1}{c|}{} 
& \multicolumn{2}{c}{\textbf{Spoon on Towel}} 
& \multicolumn{2}{c}{\textbf{Carrot on Plate}} 
& \multicolumn{2}{c}{\textbf{Stack Block}} 
& \multicolumn{2}{c}{\textbf{Eggplant in Basket}} 
& \multirow{2}{*}{\textbf{Avg}} \\
\cmidrule(lr){2-3} \cmidrule(lr){4-5} \cmidrule(lr){6-7} \cmidrule(lr){8-9} 

\textbf{Model} & Grasp & Success & Grasp & Success & Grasp & Success & Grasp & Success &  \\
\midrule
Octo \;\cite{octo2024} & 34.7 & 12.5& 52.8 & 8.3& 31.9 & 0.0& 66.7 & 43.1& 16.0 \\
OpenVLA \;\cite{kim2024openvla}& 4.1 & 0.0& 33.3 & 0.0& 12.5 & 0.0& 8.3 & 4.1 & 1.0 \\
SpatialVLA \;\cite{qu2025spatialvla}& 25.0 & 20.8& 41.7 & 20.8& 58.3 & 25.0& 79.2 & 70.8& 34.4 \\
$\pi_0$ \;\cite{black2024pi0}& 45.8 & 29.1& 25.0 & 0.0& 50.0 & 16.6& 91.6 & 62.5& 27.1 \\
$\pi_0$-FAST \;\cite{pertsch2025fast}& 62.5 & 29.1& 58.5 & 21.9& 54.0 & 10.8& 83.3 & 66.6& 32.1 \\
$\pi_{0.5}$ \;\cite{pmlr-v305-black25a} & 65.3 & 44.4 & 57.0 & 29.2 & 75.0 & 18.1 & 80.5 & 63.9 & 38.9 \\
UniVLA \;\cite{wang2025unified} & \underline{83.3} & \textbf{83.3} & \underline{74.0} &  66.7 & \textbf{95.8} & \textbf{33.3} & \textbf{100.0} & \underline{95.8} & \textbf{69.8} \\
LLaDA-VLA \;\cite{wen2025llada} & - & \underline{56.9} & - & \underline{76.3} &- & 30.6 &- & 58.3 & 55.5 \\
Disc Diff VLA \;\cite{liang2025discrete} & 70.8 &29.2&58.3&29.2&62.5&20.8&91.7&70.8& 37.5\\
\midrule
\textbf{TBD-VLA}
& \textbf{94.0} & 52.0
& \textbf{93.2} & \textbf{86.8}
& \underline{77.2} & \underline{31.2}
& \textbf{100.0} & \textbf{97.2}
& \underline{66.8} \\

\bottomrule
\end{tabular}
}
\label{fig:simpler_widowx}
\centering
\vspace{6pt}
\caption{Success rates (\%) on the SimplerEnv Google Robot benchmark. ``Drawer" includes the average score for both the opening and closing drawer tasks.}
\small
\renewcommand{\arraystretch}{1.15}
\resizebox{0.97\textwidth}{!}{%
\begin{tabular}{l|cccc|cccc}
\toprule

& \multicolumn{4}{c|}{\textbf{Visual Matching}}
& \multicolumn{4}{c}{\textbf{Variant Aggregation}} \\
\cmidrule(lr){2-5} \cmidrule(lr){6-9}

\textbf{Model}
& \textbf{Pick Can} 
& \textbf{Move Near} 
& \textbf{Drawer} 
& \textbf{Avg}
& \textbf{Pick Can} 
& \textbf{Move Near} 
& \textbf{Drawer} 
& \textbf{Avg} \\
\midrule

Octo \;\cite{octo2024}& 17.0 & 4.2 & 22.7 & 16.8& 0.6 & 3.1 & 1.1 & 1.1 \\
OpenVLA \;\cite{kim2024openvla}& 16.3 & 46.2 & 35.6 & 27.7& 54.5 & 47.7 & 17.7 & 39.8 \\
SpatialVLA \;\cite{qu2025spatialvla}& 86.0 & 77.9 & \underline{57.4} & 73.8& 88.0 & 72.7 & 41.8 & 70.7 \\
$\pi_0$ \;\cite{black2024pi0}& 72.7 & 65.3 & 38.3 & 58.8& 75.2 & 63.7 & 25.6 & 54.8 \\
$\pi_0$-FAST \;\cite{pertsch2025fast}& 75.3 & 67.5 & 42.9 & 61.9& 77.6 & 68.2 & 31.3 & 59.0 \\
InternVLA-M1 \;\cite{chen2025internvlam1}& \underline{95.3} & \textbf{90.0} & 52.5 &  \underline{79.3} & \underline{97.1} & \textbf{82.0} & \underline{72.0} & \underline{83.7} \\
\midrule
\textbf{TBD-VLA}
& \textbf{99.2} & \underline{85.0} & \textbf{88.9} & \textbf{91.0}
& \textbf{97.2} & \underline{78.3} & \textbf{83.4} & \textbf{86.3} \\

\bottomrule
\end{tabular}
}
\label{fig:simpler_google}
\vspace{-12pt}
\end{table}

\paragraph{SimplerEnv}
Table~\ref{fig:simpler_widowx} and \ref{fig:simpler_google} show the simulation performance of TBD-VLA compared to other models for SimplerEnv Widow-X and Google Robot benchmarks, respectively. In SimplerEnv Widow-X benchmark, TBD-VLA achieves the second-highest success rate at 66.8\%, falling behind only UniVLA at 69.8\%. In SimplerEnv Google Robot benchmarks, TBD-VLA outperforms the baslines with 91.0\% and 86.3\% on visually matching and variant aggregation tasks, respectively.

\subsection{Real-World Experiments} 

\paragraph{Data Collection and Tasks}
We design three real-world tabletop manipulation tasks using a Franka Research 3 robot and two RealSense D435 cameras: one for global view and the other for in-hand view. Both cameras capture 720p RGB images at 15 FPS, and the images are cropped and resized to \(256 \times 256 \times 3\). A proficient expert teleoperates the robot using a VR controller. For each task, 50 demonstrations are collected.
The proposed tasks are designed to evaluate policies under challenging real-world conditions, requiring long-horizon reasoning (``put every object on the table in the basket"), dexterity (``insert the bread into the toaster"), and reactiveness (``transfer the liquid").

\vspace{-4pt}
\paragraph{Evaluation and Baselines}
We conduct a comprehensive evaluation under out-of-distribution scenarios for each task, including a different global camera viewpoint, modified language instructions, and variations in background and lighting. With one in-distribution scenario and three perturbations scenarios, each case is rolled out 20 times. In total, each method is rolled out 240 times.
For baseline, we fine-tune the $\pi_{0.5}$ DROID checkpoint on the real-world dataset. In addition, we ablate the use of RTC for each method. For additional details on evaluation procedures, see Appendix~\ref{app:real_world_eval}.

\begin{figure*}[t]
    \centering
    \includegraphics[width=\linewidth]{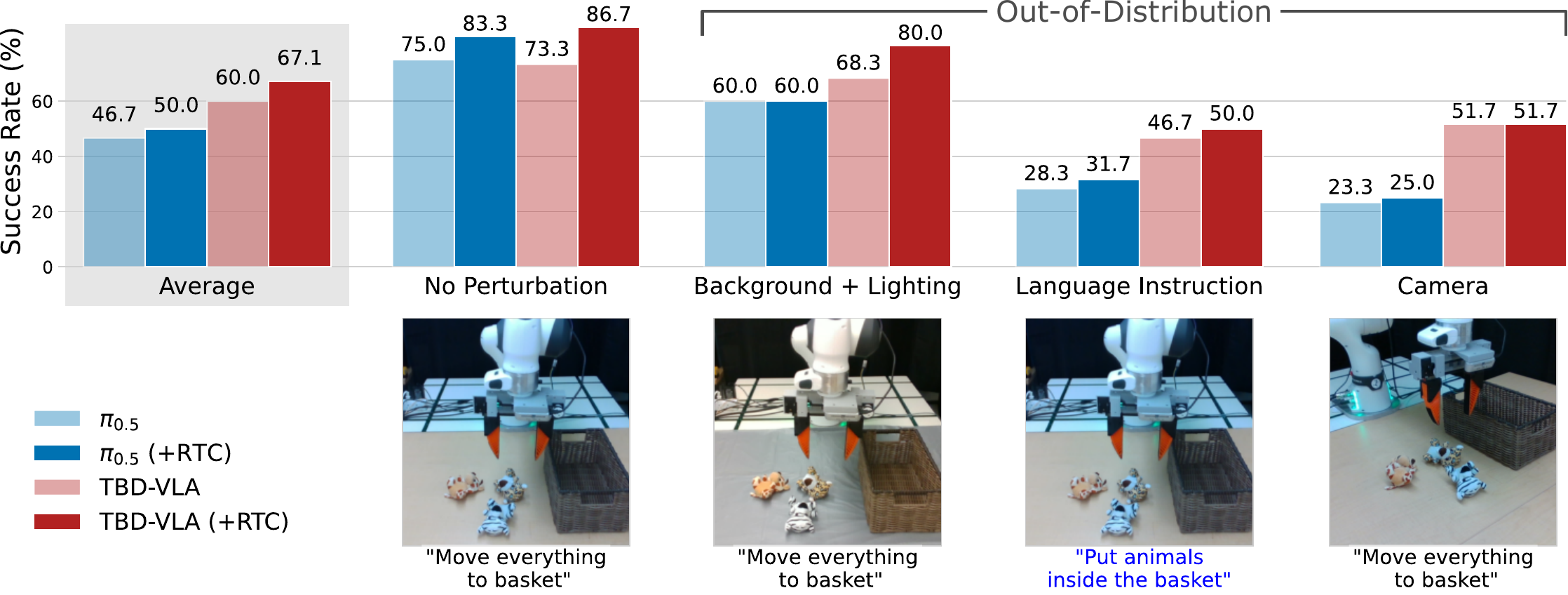}
    \vspace{-6pt}
    \caption{\textbf{Real-world evaluation results.} The average final success rate across three tasks are reported. The images represent examples of each perturbation type for ``Everything in Bin" task.}
    \label{fig:real_results}
    \vspace{-10pt}
\end{figure*}

\vspace{-4pt}
\paragraph{Results}
Figure~\ref{fig:real_results} compares TBD-VLA with $\pi_{0.5}$ on real-world tasks. Across three perturbation settings and one in-distribution setting, TBD-VLA achieves a 67.1\% average success rate over three tasks, outperforming $\pi_{0.5}$ at 50.0\%. RTC improves both methods, with TBD-VLA degrading to 60.0\% success rate without RTC. TBD-VLA maintains strong performance across out-of-distribution settings, demonstrating the effectiveness of temporal modeling with block diffusion. For more in-depth analysis of the real-world results, see Appendix~\ref{app:real_world_results}.

\begin{table}[b]
\centering
\renewcommand{\arraystretch}{1.15}
\setlength{\tabcolsep}{8pt}
\vspace{-6pt}
\caption{SimplerEnv Google Robot benchmark results comparing overall success rate, inference time, and the number of VLM forward passes across temporal block size $m$, per-block diffusion steps $n_d$, and action sampling method. The number of VLM forward passes is calculated as $\lceil H_{\mathrm{a}}/m\rceil \cdot n_d$, where $H_{\mathrm{a}}$ is 8. Inference time is measured using a single NVIDIA RTX A40 GPU.}
\vspace{3pt}
\resizebox{0.99\textwidth}{!}{%
\begin{tabular}{lccc}
\toprule
\textbf{Configuration} &
\textbf{Success Rate (\%) $\uparrow$} &
\textbf{Inference Time (s) $\downarrow$} &
\textbf{VLM Forward Passes $\downarrow$} \\
\midrule
$m$ = 1,\:\; $n_d$=2,\:\: Expectation
& 84.6 {\scriptsize\textcolor{red!70!black}{(-4.1)}} 
& 0.223 {\scriptsize\textcolor{red!70!black}{(+0.137)}} 
& 16 {\scriptsize\textcolor{red!70!black}{(+12)}} \\

$m$ = 16, $n_d$=2,\:\: Expectation
& 84.0 {\scriptsize\textcolor{red!70!black}{(-4.7)}} 
& \underline{0.061} {\scriptsize\textcolor{green!50!black}{(-0.025)}} 
& \textbf{2} {\scriptsize\textcolor{green!50!black}{(-2)}} \\

$m$ = 4,\:\; $n_d$=1,\:\: Expectation
& \underline{85.7} {\scriptsize\textcolor{red!70!black}{(-3.0)}} 
& \textbf{0.060} {\scriptsize\textcolor{green!50!black}{(-0.026)}} 
& \textbf{2} {\scriptsize\textcolor{green!50!black}{(-2)}} \\

$m$ = 4,\:\; $n_d$=2,\:\: Argmax
& 81.6 {\scriptsize\textcolor{red!70!black}{(-7.1)}} 
& 0.086 {\scriptsize\textcolor{gray}{(0.000)}} 
& 4 {\scriptsize\textcolor{gray}{(0)}} \\
\midrule
\textbf{\boldmath $m$ = 4,\: $n_d$=2,\: Expectation}
& \textbf{88.7}
& 0.086 
& 4 \\
\bottomrule
\vspace{-18pt}
\label{tab:ablate1}
\end{tabular}%
}
\centering
\renewcommand{\arraystretch}{1.15}
\setlength{\tabcolsep}{8pt}
\vspace{13pt}
\caption{Inference speed breakdown. Decode-as-needed and KV caching are TBD-VLA inference optimizations, while VLM compilation applies PyTorch compilation to the VLM forward pass.}
\vspace{3pt}
\resizebox{0.85\textwidth}{!}{%
\begin{tabular}{lcccc}
\toprule
\textbf{Components} &
\textbf{Baseline} &
\textbf{Decode as Needed} &
\textbf{KV Cache} &
\textbf{VLM Compile} \\
\midrule
Inference Speed (s)
& 0.185
& 0.125 $\downarrow$ {\scriptsize\textcolor{green!50!black}{(-0.060)}}
& 0.113 $\downarrow$ {\scriptsize\textcolor{green!50!black}{(-0.012)}}
& 0.086 $\downarrow$ {\scriptsize\textcolor{green!50!black}{(-0.027)}}\\

\bottomrule
\label{tab:ablate2}
\end{tabular}%
}

\label{tab:inference}
\end{table}

\vspace{-4pt}
\subsection{Design Choice Analysis}
\label{sec:ablation}
\vspace{-2pt}
We analyze the key design choices in TBD-VLA that affect both policy performance and inference efficiency.
As shown in Table~\ref{tab:ablate1}, we study the temporal block size \(m\), the number of diffusion refinement steps \(n_d\), and the action decoding strategy.
These factors determine how the model balances temporal dependency modeling, iterative refinement quality, and decoding accuracy.
When \(m=H_p\), the model reduces to standard discrete diffusion over the full action horizon without temporal modeling; when \(m=1\), it becomes fully temporally autoregressive, which incurs higher latency without clear performance benefits. When \(n_d=1\), the policy becomes unimodal within each block, degrading the performance at the cost of faster inference. 
Finally, we find that expectation sampling substantially improves policy performance by using the full predicted token distribution rather than choosing tokens with the maximum logits.
To best balance the policy performance and the inference latency., we use \(m=4\), \(n_d=2\), and expectation sampling as the final configuration.
Table~\ref{tab:ablate2} further breaks down the inference-time improvements from each efficiency component. Decoding only the required action blocks reduces latency from 0.185s to 0.125s, while KV caching and PyTorch VLM compilation further reduce it to 0.113s and 0.086s, respectively.

\section{Limitations}
As this work focuses on the novel adoption of block diffusion within the VLA framework, we leave the exploration of alternative training strategies, such as co-training with auxiliary VLM objectives, and their potential benefits for TBD-VLA to future work. We also leave a deeper interpretation of the VLM-only action decoding to future work, particularly how visual-language representations are transformed into executable actions. Although TBD-VLA is generally robust to perturbations, it can still fail under certain out-of-distribution conditions. For example, in the “transfer the liquid” task, a modified camera viewpoint can lead to complete failure, likely because the task requires accurate visual fidelity. Future work could improve robustness by scaling up training data and model size, as well as exploring more advanced training strategies such as co-training with auxiliary objectives.

\section{Conclusion}
\label{sec:conclusion}
We presented TBD-VLA, a discrete token-based VLA framework that combines temporal autoregression with parallel action decoding through block discrete diffusion. TBD-VLA denoises tokens within each temporal block in parallel while generating blocks autoregressively, preserving VLM-compatible action generation and explicitly modeling temporal dependencies. Across simulated and real-world manipulation tasks, TBD-VLA achieves strong generalization, robustness, and competitive latency, while compatible with Real-Time Chunking. These results highlight temporal block diffusion as a promising direction for temporally aware, low-latency, discrete VLA models.

\section*{Acknowledgment}
This research was partly supported by Delta Electronics Inc., Toyota Research Institute, and NSF CMMI-2443076.
We acknowledge Research Computing at the University of Virginia for providing the computational resources that made the results in this work possible.




\clearpage
\bibliography{bibliography} 

@inproceedings{arriola2025block,
  title={Block diffusion: Interpolating between autoregressive and diffusion language models},
  author={Arriola, Marianne and Gokaslan, Aaron and Chiu, Justin and Yang, Zhihan and Qi, Zhixuan and Han, Jiaqi and Sahoo, Subham and Kuleshov, Volodymyr},
  booktitle={International Conference on Learning Representations},
  volume={2025},
  pages={50726--50753},
  year={2025}
}

@inproceedings{wu2025fast,
  title     = {{Fast-dLLM} v2: Efficient Block-Diffusion {LLM}},
  author    = {Wu, Chengyue and Zhang, Hao and Xue, Shuchen and Diao, Shizhe and Fu, Yonggan and Liu, Zhijian and Molchanov, Pavlo and Luo, Ping and Han, Song and Xie, Enze},
  booktitle = {The Fourteenth International Conference on Learning Representations},
  year      = {2026},
  url       = {https://openreview.net/forum?id=1NZ3DHF9nT}
}

@InProceedings{li2024evaluating,
  title     = {Evaluating Real-World Robot Manipulation Policies in Simulation},
  author    = {Li, Xuanlin and Hsu, Kyle and Gu, Jiayuan and Mees, Oier and Pertsch, Karl and Walke, Homer Rich and Fu, Chuyuan and Lunawat, Ishikaa and Sieh, Isabel and Kirmani, Sean and Levine, Sergey and Wu, Jiajun and Finn, Chelsea and Su, Hao and Vuong, Quan and Xiao, Ted},
  booktitle = {Proceedings of The 8th Conference on Robot Learning},
  pages     = {3705--3728},
  year      = {2025},
  editor    = {Agrawal, Pulkit and Kroemer, Oliver and Burgard, Wolfram},
  volume    = {270},
  series    = {Proceedings of Machine Learning Research},
  month     = {06--09 Nov},
  publisher = {PMLR},
  url       = {https://proceedings.mlr.press/v270/li25c.html}
}

@inproceedings{liu2023libero,
  title     = {{LIBERO}: Benchmarking Knowledge Transfer for Lifelong Robot Learning},
  author    = {Liu, Bo and Zhu, Yifeng and Gao, Chongkai and Feng, Yihao and Liu, Qiang and Zhu, Yuke and Stone, Peter},
  booktitle = {Advances in Neural Information Processing Systems},
  volume    = {36},
  pages     = {44776--44791},
  year      = {2023},
  url       = {https://proceedings.neurips.cc/paper_files/paper/2023/hash/8c3c666820ea055a77726d66fc7d447f-Abstract-Datasets_and_Benchmarks.html}
}

@inproceedings{pertsch2025fast,
  title     = {{FAST}: Efficient Action Tokenization for Vision-Language-Action Models},
  author    = {Pertsch, Karl and Stachowicz, Kyle and Ichter, Brian and Driess, Danny and Nair, Suraj and Vuong, Quan and Mees, Oier and Finn, Chelsea and Levine, Sergey},
  booktitle = {Robotics: Science and Systems},
  year      = {2025},
  url       = {https://roboticsconference.org/program/papers/12/}
}

@inproceedings{liu2026oat,
  title     = {{OAT}: Ordered Action Tokenization},
  author    = {Liu, Chaoqi and Han, Xiaoshen and Gao, Jiawei and Zhao, Yue and Chen, Haonan and Du, Yilun},
  booktitle = {Robotics: Science and Systems},
  year      = {2026},
  url       = {https://github.com/Chaoqi-LIU/oat}
}

@inproceedings{wang2025vq,
  title     = {{VQ-VLA}: Improving Vision-Language-Action Models via Scaling Vector-Quantized Action Tokenizers},
  author    = {Wang, Yating and Zhu, Haoyi and Liu, Mingyu and Yang, Jiange and Fang, Hao-Shu and He, Tong},
  booktitle = {Proceedings of the IEEE/CVF International Conference on Computer Vision},
  pages     = {11089--11099},
  year      = {2025}
}

@article{fei2025libero,
  title   = {{LIBERO-Plus}: In-depth Robustness Analysis of Vision-Language-Action Models},
  author  = {Fei, Senyu and Wang, Siyin and Shi, Junhao and Dai, Zihao and Cai, Jikun and Qian, Pengfang and Ji, Li and He, Xinzhe and Zhang, Shiduo and Fei, Zhaoye and Fu, Jinlan and Gong, Jingjing and Qiu, Xipeng},
  journal = {arXiv preprint arXiv:2510.13626},
  year    = {2025}
}

@InProceedings{walke2023bridgedata,
  title     = {{BridgeData} V2: A Dataset for Robot Learning at Scale},
  author    = {Walke, Homer Rich and Black, Kevin and Zhao, Tony Z. and Vuong, Quan and Zheng, Chongyi and Hansen-Estruch, Philippe and He, Andre Wang and Myers, Vivek and Kim, Moo Jin and Du, Max and Lee, Abraham and Fang, Kuan and Finn, Chelsea and Levine, Sergey},
  booktitle = {Proceedings of The 7th Conference on Robot Learning},
  pages     = {1723--1736},
  year      = {2023},
  editor    = {Tan, Jie and Toussaint, Marc and Darvish, Kourosh},
  volume    = {229},
  series    = {Proceedings of Machine Learning Research},
  month     = {06--09 Nov},
  publisher = {PMLR},
  url       = {https://proceedings.mlr.press/v229/walke23a.html}
}

@inproceedings{brohan2022rt,
  title     = {{RT-1}: Robotics Transformer for Real-World Control at Scale},
  author    = {Brohan, Anthony and Brown, Noah and Carbajal, Justice and Chebotar, Yevgen and Dabis, Joseph and Finn, Chelsea and Gopalakrishnan, Keerthana and Hausman, Karol and Herzog, Alex and Hsu, Jasmine and Ibarz, Julian and Ichter, Brian and Irpan, Alex and Jackson, Tomas and Jesmonth, Sally and Joshi, Nikhil J. and Julian, Ryan and Kalashnikov, Dmitry and Kuang, Yuheng and Leal, Isabel and Lee, Kuang-Huei and Levine, Sergey and Lu, Yao and Malla, Utsav and Manjunath, Deeksha and Mordatch, Igor and Nachum, Ofir and Parada, Carolina and Peralta, Jodilyn and Perez, Emily and Pertsch, Karl and Quiambao, Jornell and Rao, Kanishka and Ryoo, Michael and Salazar, Grecia and Sanketi, Pannag and Sayed, Kevin and Singh, Jaspiar and Sontakke, Sumedh and Stone, Austin and Tan, Clayton and Tran, Huong and Vanhoucke, Vincent and Vega, Steve and Vuong, Quan and Xia, Fei and Xiao, Ted and Xu, Peng and Xu, Sichun and Yu, Tianhe and Zitkovich, Brianna},
  booktitle = {Robotics: Science and Systems},
  year      = {2023},
  url       = {https://www.roboticsproceedings.org/rss19/p025.pdf}
}

@inproceedings{khazatsky2024droid,
  title     = {{DROID}: A Large-Scale In-The-Wild Robot Manipulation Dataset},
  author    = {Khazatsky, Alexander and Pertsch, Karl and Nair, Suraj and Balakrishna, Ashwin and Dasari, Sudeep and Karamcheti, Siddharth and Nasiriany, Soroush and Srirama, Mohan Kumar and Chen, Lawrence Yunliang and Ellis, Kirsty and Fagan, Peter David and Hejna, Joey and Itkina, Masha and Lepert, Marion and Ma, Yecheng Jason and Miller, Patrick Tree and Wu, Jimmy and Belkhale, Suneel and Dass, Shivin and Ha, Huy and Jain, Arhan and Lee, Abraham and Lee, Youngwoon and Memmel, Marius and Park, Sungjae and Radosavovic, Ilija and Wang, Kaiyuan and Zhan, Albert and Black, Kevin and Chi, Cheng and Hatch, Kyle Beltran and Lin, Shan and Lu, Jingpei and Mercat, Jean and Rehman, Abdul and Sanketi, Pannag R. and Sharma, Archit and Simpson, Cody and Vuong, Quan and Walke, Homer Rich and Wulfe, Blake and Xiao, Ted and Yang, Jonathan Heewon and Yavary, Arefeh and Zhao, Tony Z. and Agia, Christopher and Baijal, Rohan and Castro, Mateo Guaman and Chen, Daphne and Chen, Qiuyu and Chung, Trinity and Drake, Jaimyn and Foster, Ethan Paul and Gao, Jensen and Herrera, David Antonio and Heo, Minho and Hsu, Kyle and Hu, Jiaheng and Jackson, Donovon and Le, Charlotte and Li, Yunshuang and Lin, Kevin and Lin, Roy and Ma, Zehan and Maddukuri, Abhiram and Mirchandani, Suvir and Morton, Daniel and Nguyen, Tony and O'Neill, Abigail and Scalise, Rosario and Seale, Derick and Son, Victor and Tian, Stephen and Tran, Emi and Wang, Andrew E. and Wu, Yilin and Xie, Annie and Yang, Jingyun and Yin, Patrick and Zhang, Yunchu and Bastani, Osbert and Berseth, Glen and Bohg, Jeannette and Goldberg, Ken and Gupta, Abhinav and Gupta, Abhishek and Jayaraman, Dinesh and Lim, Joseph J. and Malik, Jitendra and Mart{\'i}n-Mart{\'i}n, Roberto and Ramamoorthy, Subramanian and Sadigh, Dorsa and Song, Shuran and Wu, Jiajun and Yip, Michael C. and Zhu, Yuke and Kollar, Thomas and Levine, Sergey and Finn, Chelsea},
  booktitle = {Robotics: Science and Systems},
  year      = {2024},
  url       = {https://roboticsconference.org/2024/program/papers/120/}
}

@inproceedings{o2024open,
  title     = {Open X-Embodiment: Robotic Learning Datasets and {RT-X} Models},
  author    = {O'Neill, Abby and Rehman, Abdul and Maddukuri, Abhiram and Gupta, Abhishek and Padalkar, Abhishek and Lee, Abraham and Pooley, Acorn and Gupta, Agrim and Mandlekar, Ajay and Jain, Ajinkya and others},
  booktitle = {2024 IEEE International Conference on Robotics and Automation},
  pages     = {6892--6903},
  year      = {2024},
  organization = {IEEE}
}

@inproceedings{kumar2023robohive,
  title     = {{RoboHive}: A Unified Framework for Robot Learning},
  author    = {Kumar, Vikash and Shah, Rutav and Zhou, Gaoyue and Moens, Vincent and Caggiano, Vittorio and Gupta, Abhishek and Rajeswaran, Aravind},
  booktitle = {Advances in Neural Information Processing Systems},
  volume    = {36},
  pages     = {44323--44340},
  year      = {2023},
  url       = {https://papers.neurips.cc/paper_files/paper/2023/hash/8a84a4341c375b8441b36836bb343d4e-Abstract-Datasets_and_Benchmarks.html}
}

@inproceedings{wu2024robomind,
  title     = {{RoboMIND}: Benchmark on Multi-embodiment Intelligence Normative Data for Robot Manipulation},
  author    = {Wu, Kun and Hou, Chengkai and Liu, Jiaming and Che, Zhengping and Ju, Xiaozhu and Yang, Zhuqin and Li, Meng and Zhao, Yinuo and Xu, Zhiyuan and Yang, Guang and others},
  booktitle = {Robotics: Science and Systems},
  year      = {2025},
  url       = {https://roboticsconference.org/program/papers/152/}
}

@inproceedings{fang2023rh20t,
  title     = {{RH20T}: A Comprehensive Robotic Dataset for Learning Diverse Skills in One-Shot},
  author    = {Fang, Hao-Shu and Fang, Hongjie and Tang, Zhenyu and Liu, Jirong and Wang, Chenxi and Wang, Junbo and Zhu, Haoyi and Lu, Cewu},
  booktitle = {2024 IEEE International Conference on Robotics and Automation},
  year      = {2024},
  organization = {IEEE},
  url       = {https://rh20t.github.io/}
}

@inproceedings{kim2025fine,
  title     = {Fine-Tuning Vision-Language-Action Models: Optimizing Speed and Success},
  author    = {Kim, Moo Jin and Finn, Chelsea and Liang, Percy},
  booktitle = {Robotics: Science and Systems},
  year      = {2025},
  url       = {https://roboticsconference.org/program/papers/22/}
}

@inproceedings{chen2025unified,
  title     = {Unified Diffusion {VLA}: Vision-Language-Action Model via Joint Discrete Denoising Diffusion Process},
  author    = {Chen, Jiayi and Song, Wenxuan and Ding, Pengxiang and Zhou, Ziyang and Zhao, Han and Tang, Feilong and Wang, Donglin and Li, Haoang},
  booktitle = {The Fourteenth International Conference on Learning Representations},
  year      = {2026},
  url       = {https://openreview.net/forum?id=a4487c0ccbdde853b9fe256554903e70db5f15e2}
}

@inproceedings{liang2025discrete,
  title     = {Discrete Diffusion {VLA}: Bringing Discrete Diffusion to Action Decoding in Vision-Language-Action Policies},
  author    = {Liang, Zhixuan and Li, Yizhuo and Yang, Tianshuo and Wu, Chengyue and Mao, Sitong and Pei, Liuao and Yang, Xiaokang and Pang, Jiangmiao and Mu, Yao and Luo, Ping},
  booktitle = {The Fourteenth International Conference on Learning Representations},
  year      = {2026},
  url       = {https://openreview.net/forum?id=YWeNCMxdhM}
}

@article{wen2025llada,
  title   = {{LLaDA-VLA}: Vision Language Diffusion Action Models},
  author  = {Wen, Yuqing and Li, Hebei and Gu, Kefan and Zhao, Yucheng and Wang, Tiancai and Sun, Xiaoyan},
  journal = {arXiv preprint arXiv:2509.06932},
  year    = {2025}
}

@inproceedings{
black2026real,
title={Real-Time Execution of Action Chunking Flow Policies},
author={Kevin Black and Manuel Y Galliker and Sergey Levine},
booktitle={The Thirty-ninth Annual Conference on Neural Information Processing Systems},
year={2026},
url={https://openreview.net/forum?id=UkR2zO5uww}
}

@article{goyal2025vla,
  title={Vla-0: Building state-of-the-art vlas with zero modification},
  author={Goyal, Ankit and Hadfield, Hugo and Yang, Xuning and Blukis, Valts and Ramos, Fabio},
  journal={arXiv preprint arXiv:2510.13054},
  year={2025}
}

@article{song2026fast,
  title={Fast-dvla: Accelerating discrete diffusion vla to real-time performance},
  author={Song, Wenxuan and Chen, Jiayi and Chen, Shuai and Wang, Jingbo and Ding, Pengxiang and Zhao, Han and Qin, Yikai and Zheng, Xinhu and Wang, Donglin and Wang, Yan and others},
  journal={arXiv preprint arXiv:2603.25661},
  year={2026}
}

@article{wang2025unified,
  title={Unified vision-language-action model},
  author={Wang, Yuqi and Li, Xinghang and Wang, Wenxuan and Zhang, Junbo and Li, Yingyan and Chen, Yuntao and Wang, Xinlong and Zhang, Zhaoxiang},
  journal={arXiv preprint arXiv:2506.19850},
  year={2025}
}

@article{lee2025molmoact,
  title={Molmoact: Action reasoning models that can reason in space},
  author={Lee, Jason and Duan, Jiafei and Fang, Haoquan and Deng, Yuquan and Liu, Shuo and Li, Boyang and Fang, Bohan and Zhang, Jieyu and Wang, Yi Ru and Lee, Sangho and others},
  journal={arXiv preprint arXiv:2508.07917},
  year={2025}
}

@inproceedings{bjorck2025gr00t,
  archivePrefix = {arxiv},
  eprint     = {2503.14734},
  title      = {{GR00T} {N1}: An Open Foundation Model for Generalist Humanoid Robots},
  author     = {NVIDIA and Johan Bjorck and Fernando Castañeda, Nikita Cherniadev and Xingye Da and Runyu Ding and Linxi "Jim" Fan and Yu Fang and Dieter Fox and Fengyuan Hu and Spencer Huang and Joel Jang and Zhenyu Jiang and Jan Kautz and Kaushil Kundalia and Lawrence Lao and Zhiqi Li and Zongyu Lin and Kevin Lin and Guilin Liu and Edith Llontop and Loic Magne and Ajay Mandlekar and Avnish Narayan and Soroush Nasiriany and Scott Reed and You Liang Tan and Guanzhi Wang and Zu Wang and Jing Wang and Qi Wang and Jiannan Xiang and Yuqi Xie and Yinzhen Xu and Zhenjia Xu and Seonghyeon Ye and Zhiding Yu and Ao Zhang and Hao Zhang and Yizhou Zhao and Ruijie Zheng and Yuke Zhu},
  month      = {March},
  year       = {2025},
  booktitle  = {ArXiv Preprint},
}

@InProceedings{pmlr-v305-black25a,
  title = 	 {$\pi_{0.5}$: a Vision-Language-Action Model with Open-World Generalization},
  author =       {Black, Kevin and Brown, Noah and Darpinian, James and Dhabalia, Karan and Driess, Danny and Esmail, Adnan and Equi, Michael Robert and Finn, Chelsea and Fusai, Niccolo and Galliker, Manuel Y. and Ghosh, Dibya and Groom, Lachy and Hausman, Karol and ichter, brian and Jakubczak, Szymon and Jones, Tim and Ke, Liyiming and LeBlanc, Devin and Levine, Sergey and Li-Bell, Adrian and Mothukuri, Mohith and Nair, Suraj and Pertsch, Karl and Ren, Allen Z. and Shi, Lucy Xiaoyang and Smith, Laura and Springenberg, Jost Tobias and Stachowicz, Kyle and Tanner, James and Vuong, Quan and Walke, Homer and Walling, Anna and Wang, Haohuan and Yu, Lili and Zhilinsky, Ury},
  booktitle = 	 {Proceedings of The 9th Conference on Robot Learning},
  pages = 	 {17--40},
  year = 	 {2025},
  editor = 	 {Lim, Joseph and Song, Shuran and Park, Hae-Won},
  volume = 	 {305},
  series = 	 {Proceedings of Machine Learning Research},
  month = 	 {27--30 Sep},
  publisher =    {PMLR},
  url = 	 {https://proceedings.mlr.press/v305/black25a.html}
}

@article{shukor2025smolvla,
  title={Smolvla: A vision-language-action model for affordable and efficient robotics},
  author={Shukor, Mustafa and Aubakirova, Dana and Capuano, Francesco and Kooijmans, Pepijn and Palma, Steven and Zouitine, Adil and Aractingi, Michel and Pascal, Caroline and Russi, Martino and Marafioti, Andres and others},
  journal={arXiv preprint arXiv:2506.01844},
  year={2025}
}

@InProceedings{kim2024openvla,
  title     = {{OpenVLA}: An Open-Source Vision-Language-Action Model},
  author    = {Kim, Moo Jin and Pertsch, Karl and Karamcheti, Siddharth and Xiao, Ted and Balakrishna, Ashwin and Nair, Suraj and Rafailov, Rafael and Foster, Ethan P. and Sanketi, Pannag R. and Vuong, Quan and Kollar, Thomas and Burchfiel, Benjamin and Tedrake, Russ and Sadigh, Dorsa and Levine, Sergey and Liang, Percy and Finn, Chelsea},
  booktitle = {Proceedings of The 8th Conference on Robot Learning},
  pages     = {2679--2713},
  year      = {2025},
  editor    = {Agrawal, Pulkit and Kroemer, Oliver and Burgard, Wolfram},
  volume    = {270},
  series    = {Proceedings of Machine Learning Research},
  month     = {06--09 Nov},
  publisher = {PMLR},
  url       = {https://proceedings.mlr.press/v270/kim25c.html}
}

@InProceedings{sohl2015deep,
  title = 	 {Deep Unsupervised Learning using Nonequilibrium Thermodynamics},
  author = 	 {Sohl-Dickstein, Jascha and Weiss, Eric and Maheswaranathan, Niru and Ganguli, Surya},
  booktitle = 	 {Proceedings of the 32nd International Conference on Machine Learning},
  pages = 	 {2256--2265},
  year = 	 {2015},
  editor = 	 {Bach, Francis and Blei, David},
  volume = 	 {37},
  series = 	 {Proceedings of Machine Learning Research},
  address = 	 {Lille, France},
  month = 	 {07--09 Jul},
  publisher =    {PMLR},
  url = 	 {https://proceedings.mlr.press/v37/sohl-dickstein15.html}
}

@article{nie2026large,
  title={Large language diffusion models},
  author={Nie, Shen and Zhu, Fengqi and You, Zebin and Zhang, Xiaolu and Ou, Jingyang and Hu, Jun and Zhou, Jun and Lin, Yankai and Wen, Ji-Rong and Li, Chongxuan},
  journal={Advances in Neural Information Processing Systems},
  volume={38},
  pages={50608--50646},
  year={2026}
}

@article{sahoo2024simple,
  title={Simple and effective masked diffusion language models},
  author={Sahoo, Subham S and Arriola, Marianne and Schiff, Yair and Gokaslan, Aaron and Marroquin, Edgar and Chiu, Justin T and Rush, Alexander and Kuleshov, Volodymyr},
  journal={Advances in Neural Information Processing Systems},
  volume={37},
  pages={130136--130184},
  year={2024}
}

@article{swerdlow2025unified,
  title={Unified multimodal discrete diffusion},
  author={Swerdlow, Alexander and Prabhudesai, Mihir and Gandhi, Siddharth and Pathak, Deepak and Fragkiadaki, Katerina},
  journal={arXiv preprint arXiv:2503.20853},
  year={2025}
}

@article{yang2026mmada,
  title={Mmada: Multimodal large diffusion language models},
  author={Yang, Ling and Tian, Ye and Li, Bowen and Zhang, Xinchen and Shen, Ke and Tong, Yunhai and Wang, Mengdi},
  journal={Advances in Neural Information Processing Systems},
  volume={38},
  pages={138867--138907},
  year={2026}
}

@article{bai2025qwen3,
  title={Qwen3-vl technical report},
  author={Bai, Shuai and Cai, Yuxuan and Chen, Ruizhe and Chen, Keqin and Chen, Xionghui and Cheng, Zesen and Deng, Lianghao and Ding, Wei and Gao, Chang and Ge, Chunjiang and others},
  journal={arXiv preprint arXiv:2511.21631},
  year={2025}
}

@inproceedings{octo2024,
  title     = {Octo: An Open-Source Generalist Robot Policy},
  author    = {Ghosh, Dibya and Walke, Homer Rich and Pertsch, Karl and Black, Kevin and Mees, Oier and Dasari, Sudeep and Hejna, Joey and Kreiman, Tobias and Xu, Charles and Luo, Jianlan and Tan, You Liang and Chen, Lawrence Yunliang and Vuong, Quan and Xiao, Ted and Sanketi, Pannag R. and Sadigh, Dorsa and Finn, Chelsea and Levine, Sergey},
  booktitle = {Robotics: Science and Systems},
  year      = {2024},
  doi       = {10.15607/RSS.2024.XX.090},
  url       = {https://www.roboticsproceedings.org/rss20/p090.pdf}
}

@inproceedings{qu2025spatialvla,
  title     = {{SpatialVLA}: Exploring Spatial Representations for Visual-Language-Action Models},
  author    = {Qu, Delin and Song, Haoming and Chen, Qizhi and Yao, Yuanqi and Ye, Xinyi and Gu, Jiayuan and Wang, Zhigang and Ding, Yan and Zhao, Bin and Wang, Dong and Li, Xuelong},
  booktitle = {Robotics: Science and Systems},
  year      = {2025},
  doi       = {10.15607/RSS.2025.XXI.011},
  url       = {https://www.roboticsproceedings.org/rss21/p011.pdf}
}

@article{black2024pi0,
  title={$\pi_0$: A Vision-Language-Action Flow Model for General Robot Control},
  author={Black, Kevin and Brown, Noah and Driess, Danny and Esmail, Adnan and Equi, Michael and Finn, Chelsea and Fusai, Niccolo and Groom, Lachy and Hausman, Karol and Ichter, Brian and Jakubczak, Szymon and Jones, Tim and Ke, Liyiming and Levine, Sergey and Li-Bell, Adrian and Mothukuri, Mohith and Nair, Suraj and Pertsch, Karl and Shi, Lucy Xiaoyang and Tanner, James and Vuong, Quan and Walling, Anna and Wang, Haohuan and Zhilinsky, Ury},
  journal={arXiv preprint arXiv:2410.24164},
  year={2024}
}

@article{chen2025internvlam1,
  title={InternVLA-M1: A Spatially Guided Vision-Language-Action Framework for Generalist Robot Policy},
  author={Chen, Xinyi and Chen, Yilun and Fu, Yanwei and Gao, Ning and Jia, Jiaya and Jin, Weiyang and Li, Hao and Mu, Yao and Pang, Jiangmiao and Qiao, Yu and Tian, Yang and Wang, Bin and Wang, Bolun and Wang, Fangjing and Wang, Hanqing and Wang, Tai and Wang, Ziqin and Wei, Xueyuan and Wu, Chao and Yang, Shuai and Ye, Jinhui and Yu, Junqiu and Zeng, Jia and Zhang, Jingjing and Zhang, Jinyu and Zhang, Shi and Zheng, Feng and Zhou, Bowen and Zhu, Yangkun},
  journal={arXiv preprint arXiv:2510.13778},
  year={2025}
}

@inproceedings{shah2023mutex,
	title        = {MUTEX: Learning Unified Policies from Multimodal Task Specifications},
	author       = {Rutav Shah and Roberto Mart{\'\i}n-Mart{\'\i}n and Yuke Zhu},
	year         = 2023,
	booktitle    = {7th Annual Conference on Robot Learning},
	url          = {https://openreview.net/forum?id=PwqiqaaEzJ}
}

@inproceedings{belkhale2023hydra,

 title={HYDRA: Hybrid Robot Actions for Imitation Learning},
 author={Belkhale, Suneel and Cui, Yuchen and Sadigh, Dorsa},
 booktitle={Proceedings of the 7th Conference on Robot Learning (CoRL)},
 year={2023}
}

@inproceedings{nasiriany2022sailor,
  title={Learning and Retrieval from Prior Data for Skill-based Imitation Learning},
  author={Soroush Nasiriany and Tian Gao and Ajay Mandlekar and Yuke Zhu},
  booktitle={Conference on Robot Learning (CoRL)},
  year={2022}
}

@misc{BerkeleyUR5Website,
  title = {Berkeley {UR5} Demonstration Dataset},
  author = {Lawrence Yunliang Chen and Simeon Adebola and Ken Goldberg},
  howpublished = {\url{https://sites.google.com/view/berkeley-ur5/home}},
}

@inproceedings{zhou2023train,
  author={Zhou, Gaoyue and Dean, Victoria and Srirama, Mohan Kumar and Rajeswaran, Aravind and Pari, Jyothish and Hatch, Kyle and Jain, Aryan and Yu, Tianhe and Abbeel, Pieter and Pinto, Lerrel and Finn, Chelsea and Gupta, Abhinav},
  booktitle={2023 IEEE International Conference on Robotics and Automation (ICRA)}, 
  title={Train Offline, Test Online: A Real Robot Learning Benchmark}, 
  year={2023},
 }

@inproceedings{chi2024universal,
	title={Universal Manipulation Interface: In-The-Wild Robot Teaching Without In-The-Wild Robots},
	author={Chi, Cheng and Xu, Zhenjia and Pan, Chuer and Cousineau, Eric and Burchfiel, Benjamin and Feng, Siyuan and Tedrake, Russ and Song, Shuran},
	booktitle={Proceedings of Robotics: Science and Systems (RSS)},
	year={2024}
}

@article{kang2026learning,
  title={Learning force-regulated manipulation with a low-cost tactile-force-controlled gripper},
  author={Kang, Xuhui and Tian, Tongxuan and Lee, Sung-Wook and Huang, Binghao and Li, Yunzhu and Kuo, Yen-Ling},
  journal={arXiv preprint arXiv:2602.10013},
  year={2026}
}

@inproceedings{jang2021bc,
title={{BC}-Z: Zero-Shot Task Generalization with Robotic Imitation Learning},
author={Eric Jang and Alex Irpan and Mohi Khansari and Daniel Kappler and Frederik Ebert and Corey Lynch and Sergey Levine and Chelsea Finn},
booktitle={5th Annual Conference on Robot Learning},
year={2021},
url={https://openreview.net/forum?id=8kbp23tSGYv}}

@misc{cadene2024lerobot,
    author = {Cadene, Remi and Alibert, Simon and Soare, Alexander and Gallouedec, Quentin and Zouitine, Adil and Palma, Steven and Kooijmans, Pepijn and Aractingi, Michel and Shukor, Mustafa and Aubakirova, Dana and Russi, Martino and Capuano, Francesco and Pascal, Caroline and Choghari, Jade and Moss, Jess and Wolf, Thomas},
    title = {LeRobot: State-of-the-art Machine Learning for Real-World Robotics in Pytorch},
    howpublished = "\url{https://github.com/huggingface/lerobot}",
    year = {2024}
}
\clearpage





\appendix

\section{Training Details}
\label{app:train}

We use the LeRobot framework~\cite{cadene2024lerobot} for TBD-VLA training and policy deployment. This provides a unified pipeline for dataset loading, pre-processing, fine-tuning, and evaluation across the simulated and real-world benchmarks considered in this work. All models are trained using 4 NVIDIA A100 GPUs. For pre-training, we use gradient accumulation to support the large effective batch size.

\subsection{Pre-training}
\label{app:pre-training}

We pre-train TBD-VLA on a large-scale mixture of robot manipulation datasets spanning multiple task domains, embodiments, and camera views. The pre-training mixture contains subsets of demonstrations from DROID \cite{khazatsky2024droid}, BC-Z \cite{jang2021bc}, RoboMind \cite{wu2024robomind}, RoboSet \cite{kumar2023robohive}, MolmoAct \cite{lee2025molmoact}, RH20T \cite{fang2023rh20t} and Open-X Embodiment datasets \cite{shah2023mutex, belkhale2023hydra, nasiriany2022sailor, zhou2023train, BerkeleyUR5Website}.
Across the pre-training mixture, we use a total of 160,268 robot demonstration episodes and 32,351,396 training samples. The resulting dataset provides broad coverage over several robot platforms.
With 80K training steps, pre-training takes approximately 1,600 GPU hours.
\begin{table}[h]
\centering
\caption{Pre-training datasets used for TBD-VLA. We report the number of robot demonstration episodes and training samples.}
\vspace{4pt}
\label{tab:pre-training_datasets}
\begin{tabular}{lccc}
\toprule
\textbf{Dataset} & \textbf{\# Episodes} & \textbf{\# Samples} & \textbf{Embodiments} \\
\midrule
DROID \cite{khazatsky2024droid} & 53,282 & 14,153,535 & Franka \\
BC-Z \cite{jang2021bc} & 39,350 & 5,471,693 & Google Robot\\
RoboMind \cite{wu2024robomind} & 30,335 & 4,710,134 & Franka, UR5e \\
RH20T \cite{fang2023rh20t} & 6,991 & 2,899,179 & Flexiv, Franka, UR5 \\
RoboSet \cite{kumar2023robohive} & 18,300 & 2,551,749 & Franka \\
MolmoAct \cite{lee2025molmoact} & 7,902 & 1,110,869 & Franka \\
UT Austin Mutex \cite{shah2023mutex} & 1,500 & 361,883 & Franka\\
Stanford Hydra \cite{belkhale2023hydra} & 570 & 358,234 & Franka\\
Austin Sailor \cite{nasiriany2022sailor} & 240 & 353,094 & Franka\\
TOTO \cite{zhou2023train} & 902 & 294,139 & Franka\\
Berkeley AutoLab UR5 \cite{BerkeleyUR5Website} & 896 & 86,887 & UR5\\

\midrule
\textbf{Total} & \textbf{160,268} & \textbf{32,351,396} &  \textbf{5 Robots} \\
\bottomrule
\end{tabular}
\end{table}

\subsection{Fine-Tuning}
\label{app:fine-tuning}
After pre-training, the policy is fine-tuned on the target datasets. For the SimplerEnv benchmark, we use the Bridge-V2 dataset~\cite{walke2023bridgedata} for Widow-X evaluation and the Fractal dataset~\cite{brohan2022rt} for Google Robot evaluation, requiring approximately 40 and 60 GPU hours, respectively. For the LIBERO and LIBERO-Plus benchmarks, the policy is fine-tuned on the original LIBERO dataset, requiring approximately 120 GPU hours.

\subsection{Hyperparameters}
We summarize the hyperparameter settings in Tables~\ref{tab:shared_hyperparam}--\ref{tab:inference_hyperparam}. Table~\ref{tab:shared_hyperparam} lists the configuration shared across all training stages. Table~\ref{tab:stage_hyperparam} reports the separate settings for pre-training and fine-tuning. Finally, Table~\ref{tab:inference_hyperparam} summarizes the inference-time configuration used for fine-tuned policies.

\begin{table}[h]
\centering
\caption{Shared hyperparameters used for both pre-training and fine-tuning TBD-VLA.}
\vspace{4pt}
\label{tab:shared_hyperparam}
\begin{tabular}{lc}
\toprule
\textbf{Hyperparameter} & \textbf{Value} \\
\midrule

Prediction horizon \(H_p\) & 16 \\
Temporal block size \(m\) & 4 \\
Diffusion steps per block \(n_d\) & 2 \\
Action bins \(N_b\) & 512 \\
State/action normalization & MinMax \\
Learning rate & 1$e-4$ \\
Optimizer & AdamW \\
Weight decay & 0.01 \\
Warmup steps & 500 \\
Learning-rate schedule & Cosine Decay\\
Training precision & Bf16 \\
\bottomrule
\end{tabular}
\end{table}

\begin{table}[h]
\centering
\caption{Stage-specific training hyperparameters for TBD-VLA. Widow-X and Google Robot denote the evaluation environments from the SimplerEnv benchmark.}
\vspace{4pt}
\label{tab:stage_hyperparam}
\resizebox{0.9\linewidth}{!}{%
\begin{tabular}{lccccc}
\toprule
\textbf{Hyperparameter} 
& \textbf{Pre-training} 
& \textbf{LIBERO} 
& \textbf{Widow-X} 
& \textbf{Google Robot} 
& \textbf{Real-world} \\
\midrule
Batch size & 1008 & 72 & 256 & 256 & 72 \\
Training steps & 80K & 80K & 20K & 40K & 20K \\
\bottomrule
\end{tabular}
}
\end{table}
\begin{table}[h!]
\centering
\caption{Inference hyperparameters for the fine-tuned TBD-VLA policies. Widow-X and Google Robot denote the evaluation environments from the SimplerEnv benchmark.}
\vspace{4pt}
\label{tab:inference_hyperparam}
\resizebox{0.9\linewidth}{!}{%
\begin{tabular}{lcccc}
\toprule
\textbf{Hyperparameter} 
& \textbf{LIBERO} 
& \textbf{Widow-X} 
& \textbf{Google Robot} 
& \textbf{Real-world} \\
\midrule
Action horizon \(H_a\) & 12 & 8 & 8 & 12 \\
Diffusion steps per block \(n_d\) & 2 & 2 & 2 & 2 \\
Action decoding & Expectation & Expectation & Expectation & Expectation \\
\bottomrule
\end{tabular}
}
\end{table}

\section{Simulation Results}
\label{app:libero_results}

\subsection{Benchmark Implementations}

\textbf{LIBERO.}
We evaluate TBD-VLA on the standard LIBERO benchmark~\cite{liu2023libero}. 
Our evaluation uses the official LIBERO codebase and task definitions,\footnote{\url{https://github.com/Lifelong-Robot-Learning/LIBERO}} 
with the LeRobot evaluation wrapper for policy rollout and logging.

\textbf{LIBERO-Plus.}
For robustness evaluation, we use LIBERO-Plus~\cite{fei2025libero}, which extends LIBERO with controlled perturbation settings including camera, robot, language, lighting, background, sensor noise, and layout variations. 
We use the official LIBERO-Plus codebase and perturbation definitions,\footnote{\url{https://github.com/sylvestf/LIBERO-plus}} 
while using the LeRobot evaluation wrapper.

\textbf{SimplerEnv (Widow-X).}
For simulated Widow-X evaluation, we use the ManiSkill2\_real2sim environments for GPU-accelerated evaluations.\footnote{\url{https://github.com/simpler-env/ManiSkill2_real2sim}} 

\textbf{SimplerEnv (Google Robot).}
For simulated Google Robot evaluation, we use the official SimplerEnv benchmark implementation.\footnote{\url{https://github.com/simpler-env/SimplerEnv}} 

\subsection{LIBERO Results under Inference Latency}

Table~\ref{tab:appendix_libero_latency} reports TBD-VLA performance on the standard LIBERO suites under increasing inference latency in environment steps. Under zero latency, TBD-VLA achieves an overall success rate of 97.7\%. As latency increases, performance without RTC degrades sharply, falling to 72.3\% at Latency $L=4$. In contrast, the benefits of RTC become more pronounced as latency increases, maintaining an overall success rate of 93.2\% at $L=4$, corresponding to a +20.9 percentage-point improvement over w/o RTC. These results suggest that temporal compensation is especially important for maintaining closed-loop control reliability under severe inference delay.

\begin{table}[h]
\centering
\caption{LIBERO results under inference latency. Success rates are reported in percentage (\%). For $L>0$, values in parentheses denote absolute changes of w/ RTC relative to w/o RTC at the same latency.}
\vspace{3pt}
\label{tab:appendix_libero_latency}
\resizebox{0.95\linewidth}{!}{%
\begin{tabular}{lccccccc}
\toprule
\multirow{2}{*}{\textbf{Suite}}
& \multirow{2}{*}{$\mathbf{L=0}$}
& \multicolumn{2}{c}{$\mathbf{L=1}$}
& \multicolumn{2}{c}{$\mathbf{L=2}$}
& \multicolumn{2}{c}{$\mathbf{L=4}$} \\
\cmidrule(lr){3-4}
\cmidrule(lr){5-6}
\cmidrule(lr){7-8}
& 
& \textbf{w/o RTC} & \textbf{w/ RTC}
& \textbf{w/o RTC} & \textbf{w/ RTC}
& \textbf{w/o RTC} & \textbf{w/ RTC} \\
\midrule
LIBERO-10
& 95.6
& 93.2 
& 93.6 {\scriptsize\textcolor{green!50!black}{(+0.4)}}
& 89.0 
& 94.4 {\scriptsize\textcolor{green!50!black}{(+5.4)}}
& 69.8 
& 92.6 {\scriptsize\textcolor{green!50!black}{(+22.8)}} \\

LIBERO-Goal
& 98.6
& 95.6 
& 96.6 {\scriptsize\textcolor{green!50!black}{(+1.0)}}
& 94.0 
& 95.4 {\scriptsize\textcolor{green!50!black}{(+1.4)}}
& 83.2 
& 90.0 {\scriptsize\textcolor{green!50!black}{(+6.8)}} \\

LIBERO-Spatial
& 97.6
& 95.6 
& 96.6 {\scriptsize\textcolor{green!50!black}{(+1.0)}}
& 91.8 
& 94.8 {\scriptsize\textcolor{green!50!black}{(+3.0)}}
& 54.2 
& 93.4 {\scriptsize\textcolor{green!50!black}{(+39.2)}} \\

LIBERO-Object
& 99.0
& 99.8
& 98.6 {\scriptsize\textcolor{red!70!black}{(-1.2)}}
& 97.2 
& 97.2 {\scriptsize\textcolor{gray}{(+0.0)}}
& 82.0 
& 96.6 {\scriptsize\textcolor{green!50!black}{(+14.6)}} \\
\midrule
\textbf{Overall}
& \textbf{97.7}
& 96.1 
& \textbf{96.4} {\scriptsize\textcolor{green!50!black}{(+0.3)}}
& 93.0 
& \textbf{95.5} {\scriptsize\textcolor{green!50!black}{(+2.5)}}
& 72.3 
& \textbf{93.2} {\scriptsize\textcolor{green!50!black}{(+20.9)}} \\
\bottomrule
\end{tabular}
}
\end{table}

\subsection{LIBERO-Plus Full Results}
\begin{figure*}[t]
    \centering
    \includegraphics[width=0.5\linewidth]{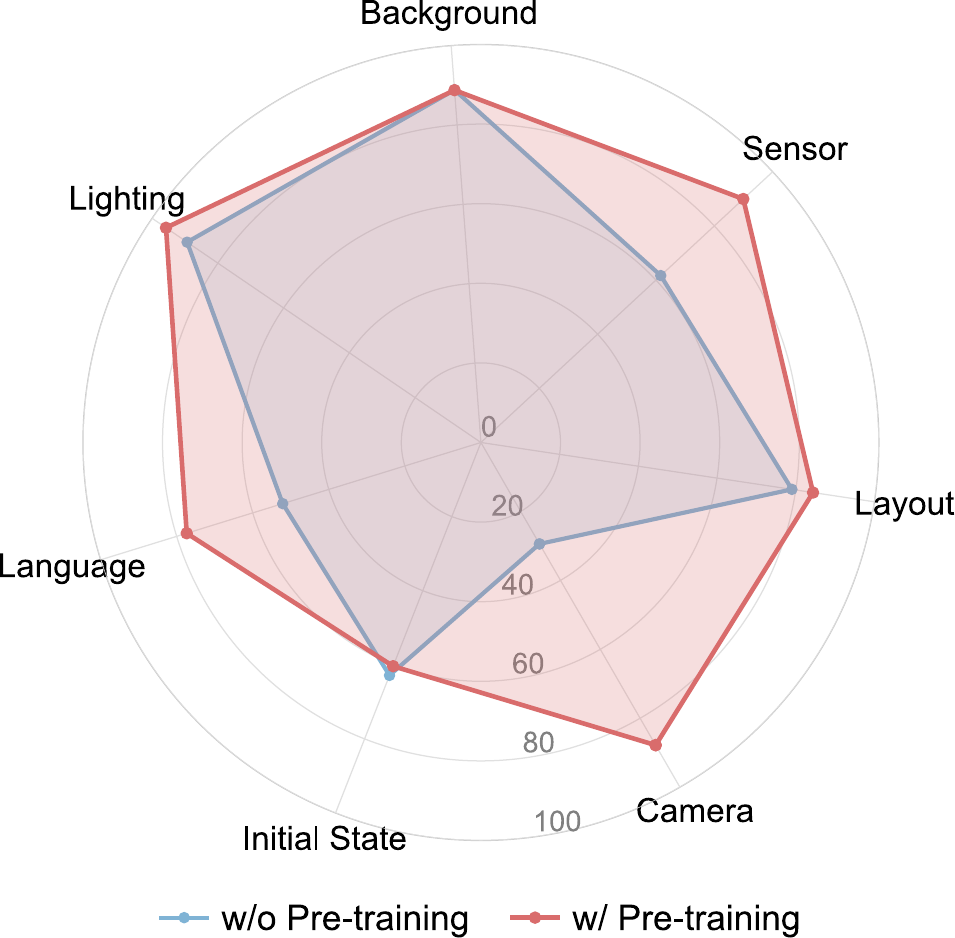}
    \vspace{-2pt}
    \caption{\textbf{Pre-training improves LIBERO-Plus robustness.} LIBERO-Plus results compared between with and without pre-training across seven perturbation settings.}
    \label{fig:pre-train}
    \vspace{0pt}
\end{figure*}

Table~\ref{tab:appendix_libero_plus} reports detailed TBD-VLA results on LIBERO-Plus under each perturbation setting. For comparison, the baseline results in Table~\ref{fig:libero_combined} are taken from the official LIBERO-Plus benchmark results~\cite{fei2025libero}.
TBD-VLA achieves an average success rate of 83.49\% across all LIBERO-Plus suites and perturbation types. 
Figure~\ref{fig:pre-train} further shows the visualization of the benefits of large-scale pre-training. Pre-training generally improves overall robustness, with larger gains under camera-viewpoint (+58.38\%), sensor-noise (+28.29\%), and language-instruction (+25.24\%) perturbations.

\begin{table}[h]
\centering
\caption{Full LIBERO-Plus robustness comparison for TBD-VLA with and without pre-training. Success rates are reported in percentage (\%). $\Delta$ indicates the improvements with pre-training}
\vspace{3pt}
\label{tab:appendix_libero_plus}
\resizebox{0.9\linewidth}{!}{%
\begin{tabular}{lcccccccc}
\toprule
\textbf{Suite} 
& \textbf{Camera}
& \textbf{Robot}
& \textbf{Language}
& \textbf{Light}
& \textbf{Background}
& \textbf{Noise}
& \textbf{Layout}
& \textbf{Avg} \\
\midrule
\multicolumn{9}{c}{\textbf{TBD-VLA w/o Pre-training}} \\
\midrule
Spatial & 31.64 & 62.28 & 52.30 & 98.97 & 94.57 & 72.36 & 94.03 & 72.31 \\
Object  & 43.43 & 65.32 & 55.93 & 98.65 & 97.17 & 63.50 & 76.42 & 71.49 \\
Goal    & 15.93 & 64.30 & 44.14 & 83.15 & 72.95 & 51.71 & 58.11 & 55.76 \\
Long    & 26.73 & 59.54 & 56.13 & 76.64 & 90.65 & 59.02 & 87.50 & 65.17 \\
\midrule
\textbf{Avg} 
& \textbf{29.43}
& \textbf{62.86}
& \textbf{52.12}
& \textbf{89.35}
& \textbf{88.84}
& \textbf{61.65}
& \textbf{79.02}
& \textbf{66.18} \\
\midrule
\multicolumn{9}{c}{\textbf{TBD-VLA w/ Pre-training}} \\
\midrule
Spatial & 99.20 & 62.57 & 78.20 & 98.28 & 95.34 & 95.15 & 97.14 & 89.41 \\
Object  & 93.69 & 69.60 & 91.24 & 99.66 & 89.52 & 98.82 & 83.87 & 89.49 \\
Goal    & 76.71 & 58.19 & 65.60 & 94.26 & 86.83 & 70.45 & 66.82 & 74.12 \\
Long    & 81.62 & 51.14 & 74.41 & 90.87 & 83.39 & 95.32 & 89.74 & 80.93 \\
\midrule
\textbf{Avg} 
& \textbf{87.81} 
& \textbf{60.38} 
& \textbf{77.36} 
& \textbf{95.77} 
& \textbf{88.77} 
& \textbf{89.94} 
& \textbf{84.39} 
& \textbf{83.49} \\
\:\:$\Delta$ 
& {\scriptsize\textcolor{green!50!black}{(+58.38)}} 
& {\scriptsize\textcolor{red!70!black}{(-2.48)}} 
& {\scriptsize\textcolor{green!50!black}{(+25.24)}} 
& {\scriptsize\textcolor{green!50!black}{(+6.42)}} 
& {\scriptsize\textcolor{red!70!black}{(-0.07)}} 
& {\scriptsize\textcolor{green!50!black}{(+28.29)}} 
& {\scriptsize\textcolor{green!50!black}{(+5.37)}} 
& {\scriptsize\textcolor{green!50!black}{(+17.31)}} \\
\bottomrule
\end{tabular}
}
\end{table}

\section{Real-World Evaluation}
\label{app:real_world_eval}

\subsection{Robot Setup}
\begin{figure*}[b!]
    \centering
    \includegraphics[width=0.7\linewidth]{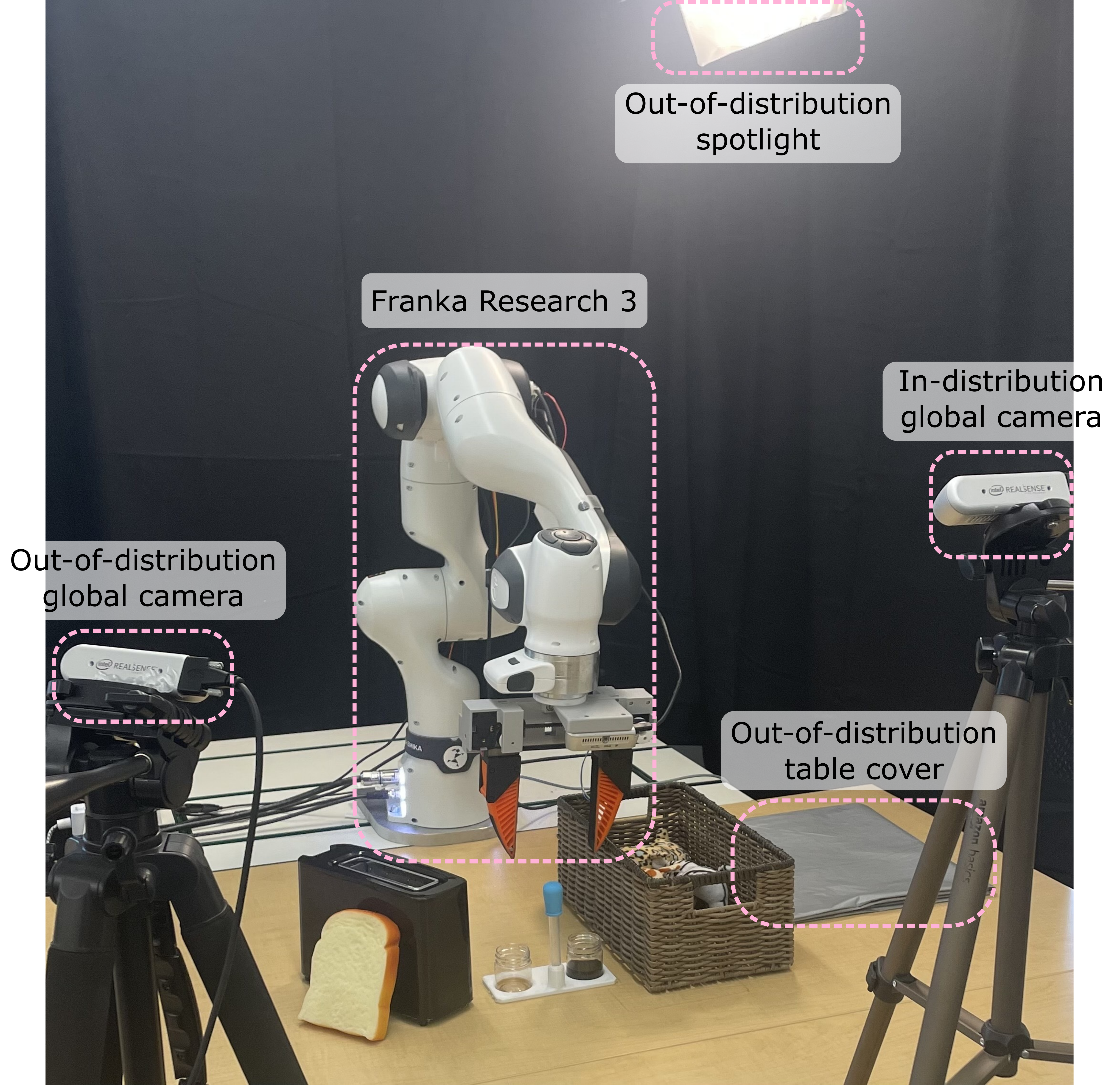}
    \vspace{-2pt}
    \caption{\textbf{Real-World Experimental Setup.} We use a Franka Research 3 robot with UMI grippers \cite{chi2024universal} for real-world manipulation. We control the gripper using width commands \cite{kang2026learning}, which are needed for precise manipulation in the ``Transfer the Liquid'' task.}
    \label{fig:real_setup}
\end{figure*}

Real-world experiments are conducted using a Franka Research 3 robot arm with two Intel RealSense D435 RGB cameras. One camera provides a global third-person view, while the other provides an in-hand view. See Figure \ref{fig:real_setup} for visualization of the real-world experiment setup. 

\subsection{Task Descriptions and Success Condition}
For real-world experiments, we evaluate TBD-VLA on the three following tabletop manipulation tasks:

\paragraph{Everything in Bin.}
The robot must place all three animal-shaped dolls on the table into a basket. The initial object locations and the order in which the dolls are picked and placed are randomized. Binary success is determined by whether all three dolls are successfully placed inside the basket.

\paragraph{Bread in Toaster.}
The robot must insert a bread object into the toaster. The locations of both the bread and the toaster are randomized. Binary success is determined by whether the bread is fully inserted into the toaster slot.

\paragraph{Transfer Liquid.}
The robot must pick up a small dropper, draw Coke from the container on the right, and dispense it into the container on the left. The success condition is when the liquid is successfully transferred without spilling.

\subsection{Evaluation Protocol}
Each method is evaluated under one in-distribution setting and three out-of-distribution perturbation settings: camera viewpoint, language instruction, and background/lighting. 
For each task and setting, we run 20 rollouts.
For the background/lighting perturbation, we add a gray table cover and a spotlight at the same time to introduce a visual shift. For the language perturbation, we replace the original task instructions from the three tasks, where the instructions are changed from ``move every object on the table to the basket,'' ``put the bread into the toaster,'' and ``transfer the liquid,'' to ``put animals inside the basket,'' ``load the toaster,'' and ``transfer the Coke,'' respectively. For the camera perturbation, we replace the original global-view camera with a secondary camera positioned to its left. 
For real-time chunking, we set the compensation timestep to 2, based on the measured inference latency of 0.119 seconds: At an evaluation frequency of 15 FPS, this latency corresponds to approximately 1.78 control timesteps, which we round to 2 for compensation.

\begin{figure*}[t]
    \centering
    \includegraphics[width=0.9\linewidth]{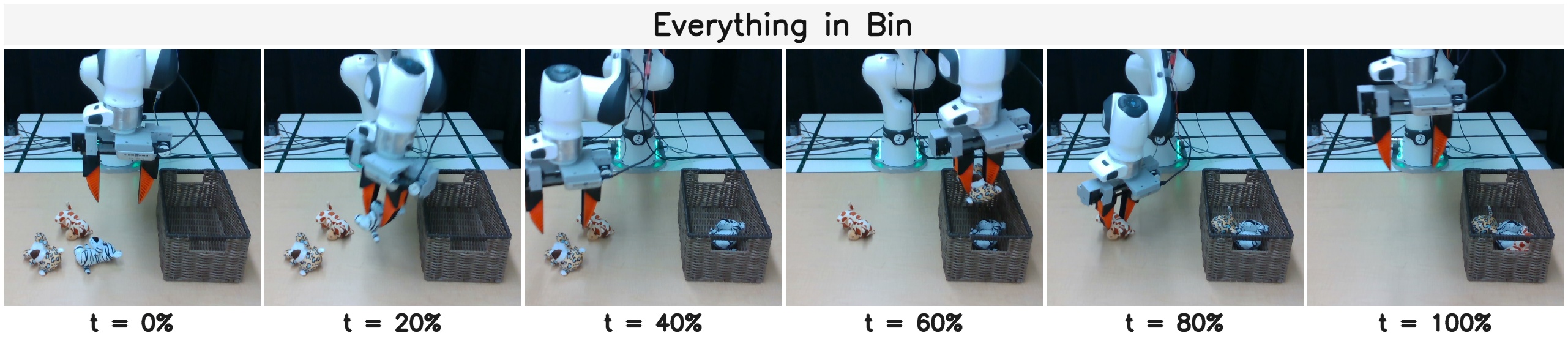}
    \includegraphics[width=0.9\linewidth]{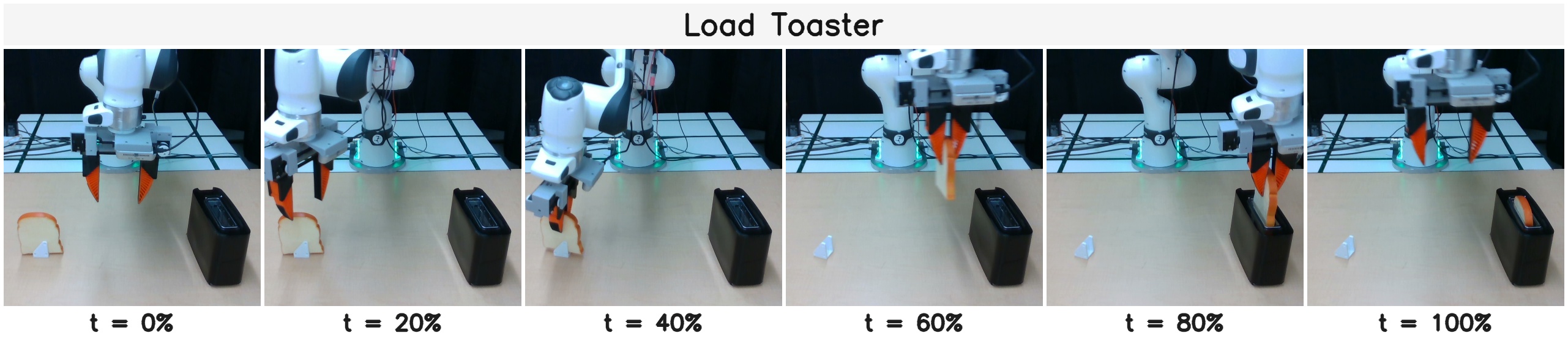}
    \includegraphics[width=0.9\linewidth]{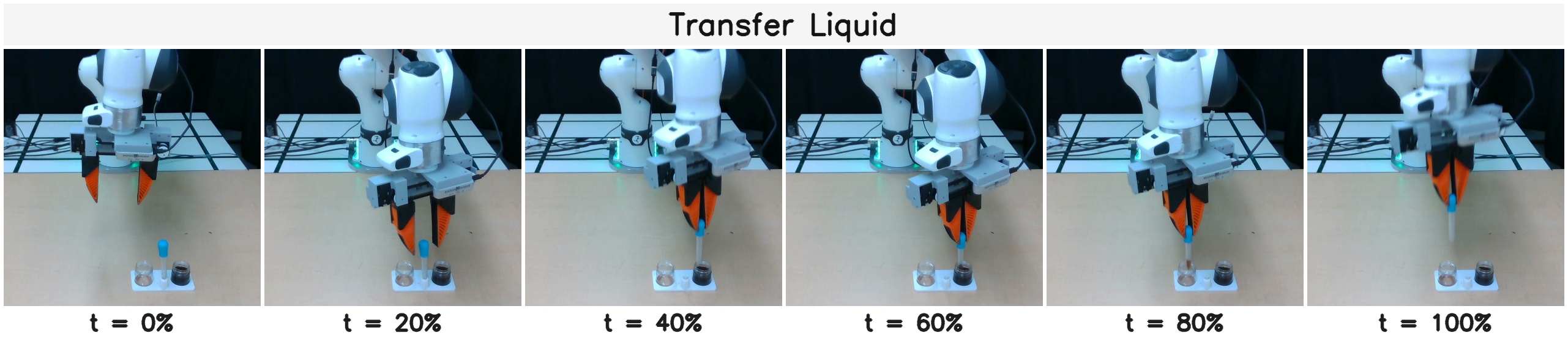}
    \caption{\textbf{Visualization of Real-world Task Progress.} For each task, the task progress is visualized at uniform time intervals during data collection.}
    \label{fig:progress}
\end{figure*}

\begin{table}[h]
\centering
\caption{Real-world success counts out of total rollouts for each task and perturbation setting. The average success rates are reported in percentage.}
\label{tab:appendix_real_task_results}
\resizebox{0.87\linewidth}{!}{%
\begin{tabular}{lccccc}
\toprule
\multicolumn{6}{c}{\textbf{Everything in Bin}} \\
\midrule
\textbf{Method} & \textbf{ID} & \textbf{Camera} & \textbf{Language} & \textbf{Background} & \textbf{Avg (\%)} \\
\midrule
\(\pi_{0.5}\) w/o RTC & 16/20 & 14/20 & 8/20  & 11/20 & 61.25 \\
\(\pi_{0.5}\) w/ RTC  & 16/20 & 15/20 & 8/20  & 12/20 & 63.75 \\
TBD-VLA w/o RTC       & 14/20 & 12/20 & 7/20  & 12/20 & 56.25 \\
TBD-VLA w/ RTC        & 17/20 & 12/20 & 6/20  & 14/20 & 61.25 \\
\midrule
\multicolumn{6}{c}{\textbf{Bread in Toaster}} \\
\midrule
\textbf{Method} & \textbf{ID} & \textbf{Camera} & \textbf{Language} & \textbf{Background} & \textbf{Avg (\%)} \\
\midrule
\(\pi_{0.5}\) w/o RTC & 17/20 & 0/20  & 9/20  & 13/20 & 48.75 \\
\(\pi_{0.5}\) w/ RTC  & 18/20 & 0/20  & 11/20 & 11/20 & 50.00 \\
TBD-VLA w/o RTC       & 17/20 & 19/20 & 13/20 & 17/20 & 82.50 \\
TBD-VLA w/ RTC        & 19/20 & 19/20 & 12/20 & 18/20 & 85.00 \\
\midrule
\multicolumn{6}{c}{\textbf{Transfer the Liquid}} \\
\midrule
\textbf{Method} & \textbf{ID} & \textbf{Camera} & \textbf{Language} & \textbf{Background} & \textbf{Avg (\%)} \\
\midrule
\(\pi_{0.5}\) w/o RTC & 12/20 & 0/20 & 0/20  & 12/20 & 30.00 \\
\(\pi_{0.5}\) w/ RTC  & 16/20 & 0/20 & 0/20  & 13/20 & 36.25 \\
TBD-VLA w/o RTC       & 13/20 & 0/20 & 8/20  & 12/20 & 41.25 \\
TBD-VLA w/ RTC        & 16/20 & 0/20 & 12/20 & 16/20 & 55.00 \\
\midrule
\textbf{TBD-VLA w/ RTC Avg (\%)} & \textbf{86.67} & \textbf{51.67} & \textbf{50.00} & \textbf{80.00} & \textbf{67.08} \\
\bottomrule
\end{tabular}
}
\end{table}

\section{Real-World Results}
\label{app:real_world_results}

We report success counts over total rollouts and average success rates for TBD-VLA and $\pi_{0.5}$ in Table~\ref{tab:appendix_real_task_results}. In the in-distribution setting, where the camera view, language instruction, and background match the training data, TBD-VLA achieves an 86.67\% success rate across the three tasks. Under the modified global camera view, modified language instructions, and background visual shift, TBD-VLA achieves success rates of 51.67\%, 50.00\%, and 80.00\%, respectively, demonstrating robustness across diverse real-world perturbations. Enabling RTC improves the average success rate by 7.08\%, showing that temporal modeling with asynchronous inference provides practical benefits in real-world settings.

\begin{figure}[h]
    \centering
    \includegraphics[width=\linewidth]{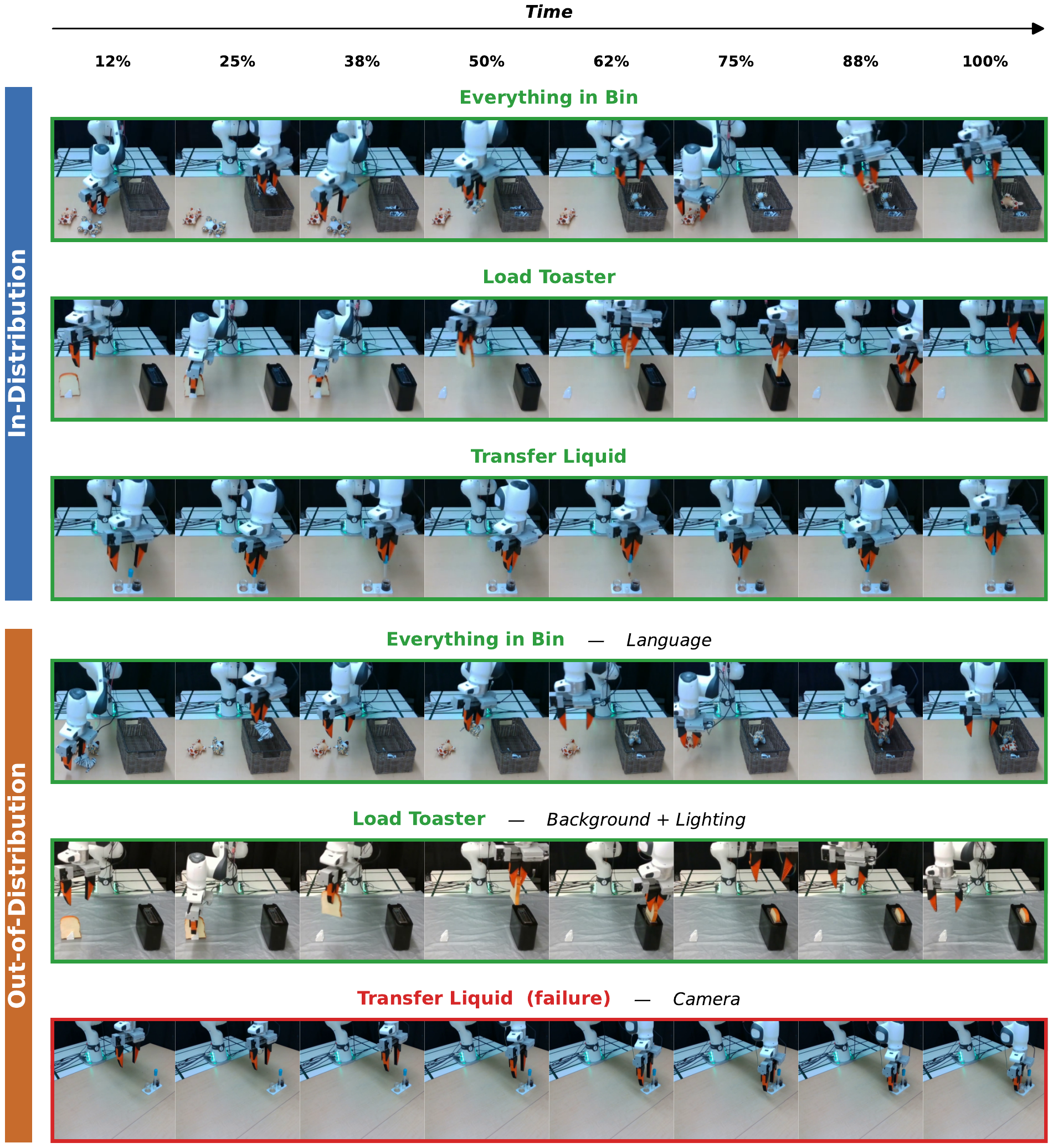}
    \caption{\textbf{Qualitative examples of real-world rollouts for both in-distribution and out-of-distribution evaluations.} We show the failure mode of TBD-VLA under camera viewpoint shift for the ``Transfer the Liquid'' task.}
    \label{fig:appendix_real_qualitative}
\end{figure}

In Figure \ref{fig:appendix_real_qualitative}, we include qualitative examples of successful and failed rollouts. TBD-VLA generally exhibits strong temporal consistency under various forms of perturbations. It is noted that with the modified camera viewpoint, TBD-VLA achieves zero success rate on ``Transfer the Liquid'' task, where the robot is unable to approach the dropper, likely due to the task's requirement for visual consistency and under-representation of similar types of tasks in the pre-training dataset.

\end{document}